\documentclass[letterpaper]{article} 
\usepackage{aaai23}  
\usepackage{times}  
\usepackage{helvet}  
\usepackage{courier}  
\usepackage[hyphens]{url}  
\usepackage{graphicx} 
\usepackage{natbib}  
\usepackage{caption} 
\usepackage{algorithm}
\usepackage{algorithmic}

\usepackage{microtype}
\usepackage{graphicx}
\usepackage{booktabs}

\usepackage{amsmath,amsfonts,bm}

\def\eqref#1{equation~\ref{#1}}

\def\1{\bm{1}}

\def\vu{{\bm{u}}}
\def\vv{{\bm{v}}}

\def\vx{{\bm{x}}}
\def\vy{{\bm{y}}}
\def\vz{{\bm{z}}}

\def\mF{{\bm{F}}}
\def\mG{{\bm{G}}}

\def\mT{{\bm{T}}}

\DeclareMathAlphabet{\mathsfit}{\encodingdefault}{\sfdefault}{m}{sl}
\SetMathAlphabet{\mathsfit}{bold}{\encodingdefault}{\sfdefault}{bx}{n}

\def\gH{{\mathcal{H}}}
\def\gI{{\mathcal{I}}}

\def\gL{{\mathcal{L}}}

\usepackage{amsmath}
\usepackage{amssymb}
\usepackage{mathtools}
\usepackage{amsthm}

\usepackage{CJK}            
\usepackage{color, colortbl}
\usepackage{array}
\usepackage{tablefootnote}
\usepackage{multirow}
\usepackage{subcaption}
\usepackage{bbm}
\usepackage{wrapfig}
\usepackage{tablefootnote}
\usepackage{threeparttable}
\usepackage{mathtools}
\usepackage{caption}
\usepackage{multirow}
\usepackage{enumitem}
\usepackage{graphicx} %
\usepackage{makecell}
\usepackage{enumitem}
\usepackage{xcolor}
\definecolor{navyblue}{rgb}{0.0, 0.0, 0.6}
\usepackage{pifont}
\newcommand{\cmark}{\ding{51}}%
\newcommand{\xmark}{\ding{55}}%

\definecolor{lightblue}{HTML}{d7eefd}
\definecolor{lightblue}{HTML}{f5e0d6}

\definecolor{anchor}{RGB}{209,123,26}
\definecolor{pos}{RGB}{0,122,0}
\definecolor{neg}{RGB}{255,51,51}

\usepackage{newfloat}
\usepackage{listings}
\DeclareCaptionStyle{ruled}{labelfont=normalfont,labelsep=colon,strut=off} 
\floatstyle{ruled}
\newfloat{listing}{tb}{lst}{}
\floatname{listing}{Listing}
\newtheorem{theorem}{Theorem}[section]
\newtheorem{proposition}[theorem]{Proposition}

\title{Self-Contrastive Learning:\\Single-viewed Supervised Contrastive Framework using Sub-network}
\author{
    Sangmin Bae\textsuperscript{\rm 1}\equalcontrib,
    Sungnyun Kim\textsuperscript{\rm 1}\equalcontrib,
    Jongwoo Ko\textsuperscript{\rm 1},
    Gihun Lee\textsuperscript{\rm 1},
    Seungjong Noh\textsuperscript{\rm 2},
    Se-Young Yun\textsuperscript{\rm 1}
}
\affiliations{
    \textsuperscript{\rm 1}Graduate School of Artificial Intelligence, KAIST\,\,
    \textsuperscript{\rm 2}SK Hynix\\
    \{bsmn0223, ksn4397, jongwoo.ko, opcrisis, yunseyoung\}@kaist.ac.kr, seungjong.noh@sk.com
}

\usepackage{bibentry}
\usepackage{hyperref}
\hypersetup{pdfborder={0 0 0.6}}

\begin{document}

\maketitle

\begin{abstract}

Contrastive loss has significantly improved performance in supervised classification tasks by using a multi-viewed framework that leverages augmentation and label information.
The augmentation enables contrast with another view of a single image but enlarges training time and memory usage.
To exploit the strength of multi-views while avoiding the high computation cost, we introduce a multi-exit architecture that outputs multiple features of a single image in a single-viewed framework.
To this end, we propose Self-Contrastive (SelfCon) learning, which self-contrasts within multiple outputs from the different levels of a single network.
The multi-exit architecture efficiently replaces multi-augmented images and leverages various information from different layers of a network.
We demonstrate that SelfCon learning improves the classification performance of the encoder network, and empirically analyze its advantages in terms of the single-view and the sub-network.
Furthermore, we provide theoretical evidence of the performance increase based on the mutual information bound.
For ImageNet classification on ResNet-50, SelfCon improves accuracy by +0.6\% with 59\% memory and 48\% time of Supervised Contrastive learning, and a simple ensemble of multi-exit outputs boosts performance up to +1.5\%.
Our code is available at \url{https://github.com/raymin0223/self-contrastive-learning}.
\end{abstract}
\section{Introduction}
\label{sec:introduction}

\begin{figure}[t]
\vspace{15pt}
    \begin{subfigure}[b]{0.54\linewidth}
    \centering
    \includegraphics[width=\linewidth]{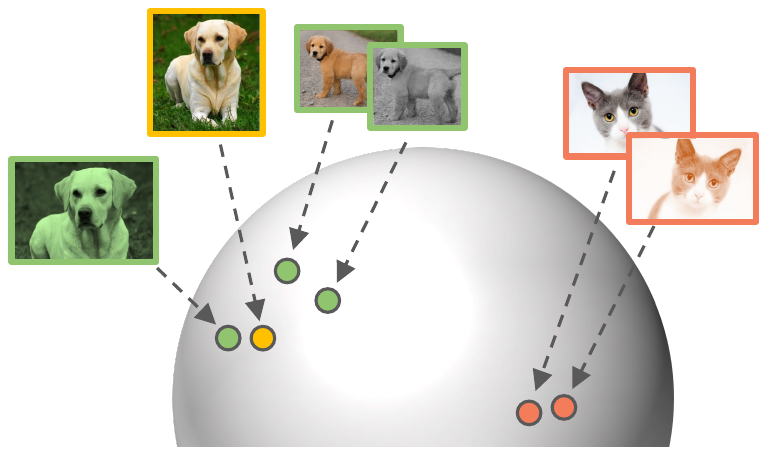}
    \caption{\small {SupCon}}\label{fig:motivation_a}
    \end{subfigure}%
    \hfill
    \begin{subfigure}[b]{0.44\linewidth}
    \centering
    \includegraphics[width=\linewidth]{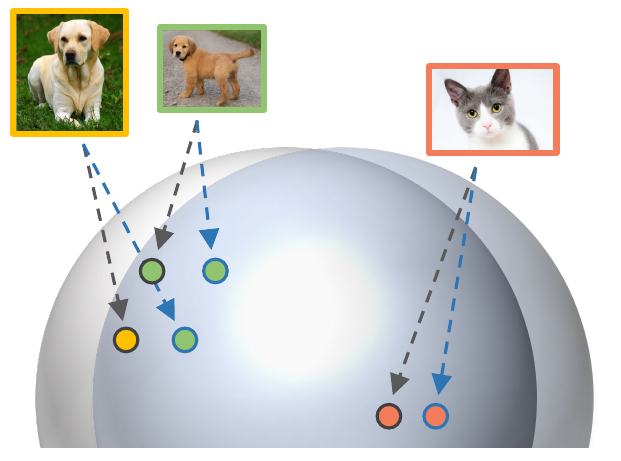}
    \caption{\small {SelfCon}}\label{fig:motivation_b}
    \end{subfigure}
    \caption{Overview of (a) SupCon learning and (b) SelfCon learning. The anchor, positive\,(which is desired to be close to the anchor), and negative\,(which is desired to be far from the anchor) samples are represented on the feature space as yellow, green, and red points, respectively. While SupCon relies on the augmentation-based multi-views, SelfCon is a single-viewed supervised contrastive learning framework. SelfCon produces multiple features from a single instance, using the sub-network.}
    \label{fig:motivation}
\end{figure}

While the cross-entropy\,(CE) loss is the most common and powerful loss function for supervised classification tasks, lots of alternatives have been proposed to overcome the shortcomings of cross-entropy, such as high generalization error \citep{large_margin, large_margin2}.
Among the various approaches, Supervised Contrastive (SupCon \citep{supcon}) loss recently showed remarkable improvement in performance for large-scale benchmarks like ImageNet \citep{imagenet}. The main idea of this loss is to make representations from the same class closer together and representations from different classes farther apart (see Figure \ref{fig:motivation_a}). 

SupCon and its related works \citep{graf2021dissecting, zheng2021weakly, chen2022perfectly, li2022targeted} have been developed on the top of a \emph{multi-viewed} framework that leverages two core factors, augmentation and label information, when formulating the contrastive task. 
Additional augmented images improve the performance by enabling contrast within a single image. The \emph{multi-viewed} framework is crucial. We indeed empirically observed that a simple extension to a \emph{single-viewed} framework (i.e., only exploiting the label information) significantly degrades the performance on large-scale datasets (Section \ref{exp:representation}). However, the augmentation-based multi-view approach makes the training time and memory usage highly expensive \citep{simclr, moco, swav}.


To implement the multi-view framework without data augmentations, we propose \textbf{Self-Contrastive (SelfCon) learning}, which uses the multi-exit architecture \citep{branchynet, be_your_own, scan, phuong2019distillation} having sub-networks that produce multiple features of a single image.
With the multi-exit architecture, SelfCon \textit{self-contrasts} within multiple outputs from the different levels of a single network (see Figure \ref{fig:motivation_b}), making the \textit{single-viewed} framework usable. Therefore, the multi-exit architecture efficiently replaces data augmentation by leveraging various information from different layers of a network \citep{visualizing_grad}.

We summarize the contributions of our paper as follows:

\noindent\textbf{[Section \ref{sec:self-contrastive_learning}]} We propose Self-Contrastive learning, which is the first study on a single-viewed contrastive framework exploiting multiple features from different levels of a single network.

\noindent \textbf{[Section \ref{sec:discussions}]} We guarantee that SelfCon loss is the lower bound of label-conditional mutual information\,(MI) between the intermediate and the last features. 
To our knowledge, this is the first work to provide the MI bound for supervised contrastive learning.

\noindent\textbf{[Section \ref{exp:representation}]} SelfCon learning efficiently achieves higher classification accuracy for various benchmarks compared to CE and SupCon loss.
Furthermore, SelfCon with an ensemble prediction boosts performance by a large margin.

\noindent\textbf{[Section \ref{exp:single_view}--\ref{exp:mi_estimation}]} We extensively investigate the benefits of SelfCon learning in terms of the single-viewed batch and the sub-network. 
Also, our empirical study of MI estimation provides evidence for the superior performance.

\section{Related Works}
\label{sec:related_work}

\subsection{Contrastive Learning in Supervision}


After \citet{cpc} proposed InfoNCE loss (also called a contrastive loss), contrastive learning-based algorithms began to show a remarkable improvement in image representation learning \citep{simclr, moco, byol, swav, simsiam}. 
\citet{supcon} extended the contrastive learning to a supervised classification task to resolve the generalization issue of cross-entropy loss. 
The idea of SupCon \citep{supcon}, which leverages augmentation and label information on the contrastive framework, has also been extended to semantic segmentation \citep{segment_supcon} and language tasks \citep{language_supcon}. 
While SupCon loss utilizes the output features from two random augmentations, our approach contrasts the features from different network paths by introducing the multi-exit framework \citep{branchynet, be_your_own}. 
In this paper, we investigate the advantages of model-based contrastive learning within the single-viewed  framework.
Moreover, we offer the first proof of the MI bound for the supervised contrastive framework to theoretically explain how SelfCon improves the classification performance.

\subsection{Multi-exit Architectures}

As earlier layers of the deep neural network suffer from the vanishing gradient issue \citep{googlenet, resnet}, previous works have introduced a multi-exit architecture \citep{deeply_supervised, branchynet, bolukbasi2017adaptive} that attaches sub-networks on the intermediate layers.
The sub-networks have also been used to predict at any point of the network during the evaluation phase (i.e., anytime inference \citep{msdnet, yang2020resolution, ruiz2021anytime}), as well as to leverage the information from different levels of a network which leads to the performance gain \cite{visualizing_grad, be_your_own, yao2020knowledge}.
Recently, the knowledge distillation-based losses \citep{on_the_fly_ensemble, be_your_own, scan, phuong2019distillation, zhang2021self} have been proposed to effectively train the sub-network.
Motivated by these methods, we propose a novel supervised contrastive learning that self-contrasts within the multi-exit outputs.
The sub-network mitigates the vanishing gradient issue and reduces the generalization error, as well as replacing the augmentation-based multi-views.

\begin{figure*}[t]
    \centering
    \includegraphics[width=0.93\linewidth]{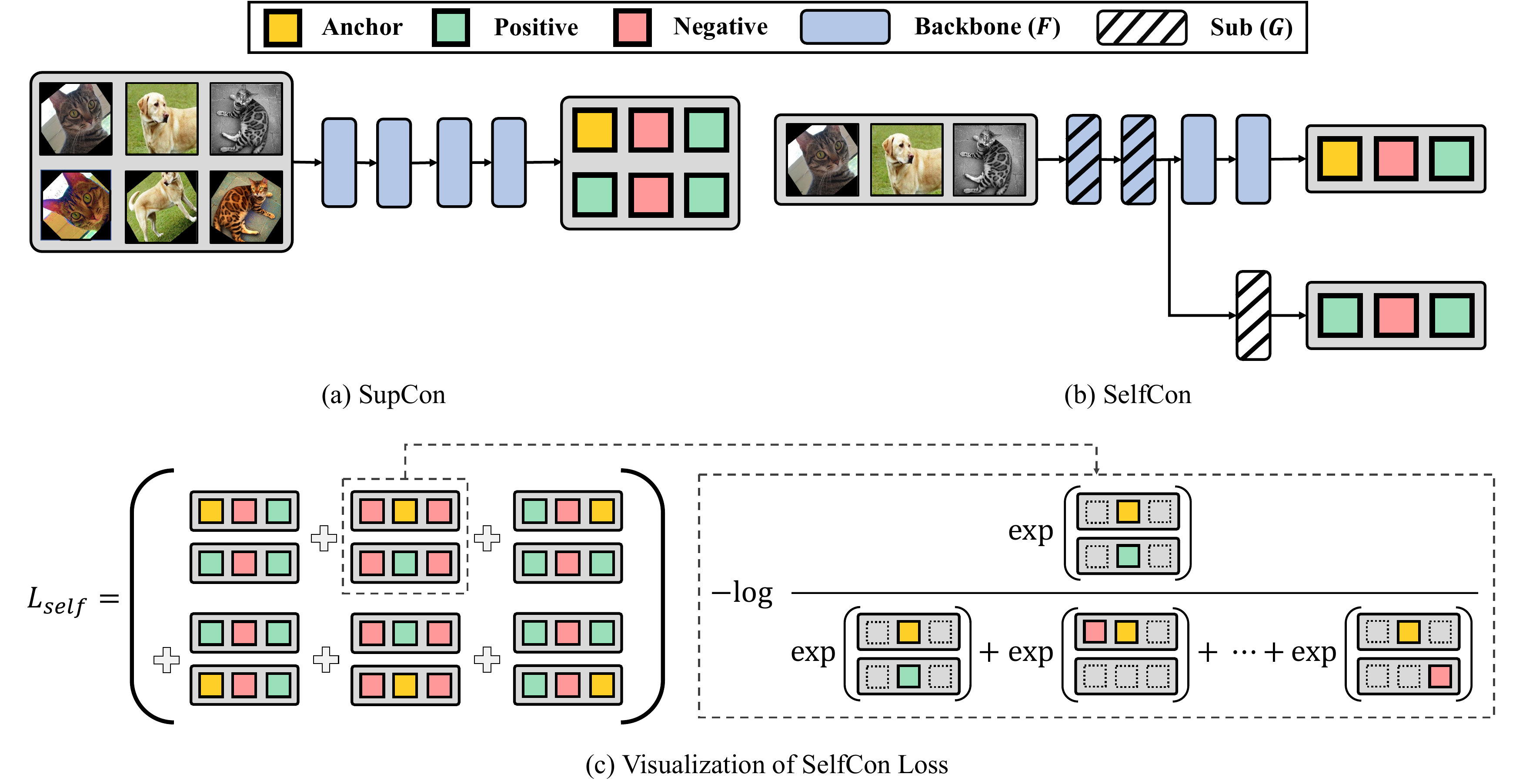}
    \caption{(Top) Comparison of learning frameworks in terms of augmentation and architecture. In both SupCon \citep{supcon} and SelfCon, every sample of the same ground-truth label with an anchor is used as a positive pair. Specifically, in SelfCon, an anchor from the backbone network contrasts other features from the backbone, as well as the features from the sub-network.
    (Bottom) We visualized the SelfCon loss function to ease the understanding in Section \ref{sec:self-contrastive_learning}. $\text{exp}(\cdot)$ denote the exponential function of the cosine similarity between two features. Note that SupCon loss has the same form but uses the representations from the multi-viewed batch.
    Best seen in color.}
    \label{fig:overview}
\end{figure*}

\section{Self-Contrastive Learning}
\label{sec:self-contrastive_learning}
\makeatletter
\newcommand*\bigcdot{\mathpalette\bigcdot@{.5}}
\newcommand*\bigcdot@[2]{\mathbin{\vcenter{\hbox{\scalebox{#2}{$\m@th#1\bullet$}}}}}
\makeatother

We propose a new supervised contrastive loss that maximizes the similarity of the outputs from different network paths by introducing the multi-exit framework.
We define an encoder structure, using $\mF$ as a backbone network and $\mG$ as a sub-network, that shares the backbone's parameters up to some intermediate layer. $\mT$ denotes the sharing layers that produce the intermediate feature. 
Note that $\mF$ and $\mG$ include the projection head after the encoder. We highlight the \textcolor{pos}{\textbf{positive}} and \textcolor{neg}{\textbf{negative}} pairs with respect to an \textcolor{anchor}{\textbf{anchor}} sample, following Figure \ref{fig:overview}.

\paragraph{SupCon loss}

To mitigate the weaknesses of cross-entropy, such as the reduced generalization performance and the possibility of poor margins, \citet{supcon} propose a supervised version of contrastive loss that defines the positive pairs as every sample with the same ground-truth label. We reformulate the SupCon loss function as follows:


\vspace{-5pt}
\begin{align}\label{eq:supcon}
    &\mathcal{L}_{sup} = \mathop{\textstyle \sum}_{\textcolor{anchor}{i} \in I} \Big[ - \tfrac{1}{|P_i|} \mathop{\textstyle \sum}_{ \textcolor{pos}{p} \in P_{i}}  \mF(\textcolor{anchor}{\vx_i}) ^\top \mF(\textcolor{pos}{\vx_{p}}) \\
    &\hspace{20pt} + \text{log} \Big(\mathop{\textstyle \sum}_{\textcolor{pos}{p} \in P_{i}} e^{\mF(\textcolor{anchor}{\vx_i}) ^\top \mF(\textcolor{pos}{\vx_{p}})} + \mathop{\textstyle \sum}_{\textcolor{neg}{n} \in N_i} e^{\mF(\textcolor{anchor}{\vx_i}) ^\top \mF(\textcolor{neg}{\vx_{n}})} \Big)  \Big] \,\,\,\,\,\, \nonumber
\end{align}
\vspace{-6pt}
\begin{align}
    & P_{i} \equiv \{ p \in I\setminus \{i\} | y_p = y_i \},\, N_i \equiv \{n\in I|y_n \neq y_i\} \nonumber
\vspace{12pt}
\end{align}
where $I \equiv \{1,\dots,2B\}$, and $B$ is the batch size. For brevity, we omit the temperature $\tau$, which softens or hardens the softmax value, and the dividing constant for the summation of anchor samples (i.e., $|I|^{-1}$). $I$ denotes a set of indices for the multi-viewed batch that concatenates the original $B$ images and the augmented ones, i.e., $\vx_{B+i}$ is an augmented pair of $\vx_i$. $P_{i}$ and $N_i$ are sets of positive and negative pair indices with respect to an anchor $i$.
Eq. \ref{eq:supcon} is a type of categorical cross-entropy loss; the numerator contains the positive pair, and the denominator contains both positive and negative pairs.

\paragraph{SelfCon loss}
We aim to maximize the similarity between the outputs from the backbone and the sub-network. To this end, we define \textbf{SelfCon loss, which forms a self-contrastive task for every output, including the features from the sub-network.}


\vspace{-5pt}
\begin{align}
  \label{eq:self-contrastive loss}
  &\mathcal{L}_{self} = \mathop{\textstyle \sum}_{\substack{ \textcolor{anchor}{i} \in I,\\\bm{\omega} \in \bm{\Omega}}} \Big[ - \tfrac{1}{|P_{i1}||\bm{\Omega}|} \mathop{\textstyle \sum}_{\substack{ \textcolor{pos}{p_1} \in P_{i1}, \\\bm{\omega}_1 \in \bm{\Omega}}}  \bm{\omega}(\textcolor{anchor}{\vx_i}) ^\top \bm{\omega}_1(\textcolor{pos}{\vx_{p_1}}) \\
  &\!+ \text{log}\!\mathop{\textstyle \sum}_{\bm{\omega}_2 \in \bm{\Omega}}\!\Big( \mathop{\textstyle \sum}_{\textcolor{pos}{p_2} \in P_{i2}}\!e^{\bm{\omega}(\textcolor{anchor}{\vx_i}) ^\top \bm{\omega}_2(\textcolor{pos}{\vx_{p_2}})}\!+\!\mathop{\textstyle \sum}_{\textcolor{neg}{n} \in N_i}\!e^{\bm{\omega}(\textcolor{anchor}{\vx_i}) ^\top \bm{\omega}_2(\textcolor{neg}{\vx_{n}})} \Big) \Big] \nonumber
\end{align}
\vspace{-6pt}
\begin{align}
    & P_{ij} \equiv \{ p_j \in I \setminus \{i\} | y_{p_j} = y_i \},\, N_i \equiv \{ n \in I | y_n \neq y_i \} \nonumber
    \vspace{12pt}
\end{align}
%
where $I \equiv \{1,\dots,B\}$, and $\bm{\Omega}=\{\mF,\mG\}$ is a function set of the backbone network and the sub-network.
We also omit $\tau$ and the dividing constant (i.e., $(|I||\bm{\Omega}|)^{-1}$).
$\bm{\omega}_1$ is a function that generates positive pair, and $\bm{\omega}_2$ is for generating every contrastive pair from a multi-exit network. 
We include an anchor sample to the positive set when the output feature is from a different exit path, i.e., $P_{ij}\leftarrow P_{ij} \cup \{i\}$ when $\bm{\omega} \neq \bm{\omega}_j$. 
For example, $\mG(\vx_i)$ is also a positive pair for $\mF(\vx_i)$.
Refer to Figure \ref{fig:overview} for better understanding of contrastive task formation in the SelfCon framework.

Whereas prevalent contrastive approaches \citep{supcon, simclr, byol} force a multi-viewed batch generated by data augmentation, the sub-network in SelfCon learning plays a role as the augmentation and provides an alternative view on the feature space. Therefore, without the additional augmented samples, we formulate our SelfCon loss function with a single-viewed batch.


We can further use multiple sub-networks, i.e., $\bm{\Omega}=\{\mF,\mG_1,\mG_2,\dots\}$. Appendix \ref{app:ablation_subnet} presents the classification performance of the expanded network, but there was no significant improvement from that of a single sub-network. Thus, we have efficiently used a single sub-network throughout our paper.

\section{Discussions}
\label{sec:discussions}


In this Section, we discuss theoretical evidence for the success of SelfCon learning. We summarize the discussion as follows: Selfcon learning improves the classification performance by encouraging the intermediate feature to have more label information in the last feature.


\medskip
\paragraph{Discussion 4.1. How does SelfCon loss encourage the intermediate feature to learn the label information in the last feature?}
Generally, prior works \citep{cpc, infomax} support the success of unsupervised contrastive learning from the connection to the MI. In this sense, in Proposition \ref{prop:supcon_loss}, we first prove the connection between a \textit{supervised} contrastive loss and the MI of positive pair features. In Proposition \ref{prop:selfcon_loss}, we then provide the MI bound within a single-viewed batch using the sub-network feature. 


\begin{proposition}\label{prop:supcon_loss}
Let $\vx$ and $\vz$ be different samples that share the same class label $c$. Then, with some discriminator function modeled by a neural network $\mF$ and $2(K-1)$ negative sample size, SupCon loss maximizes the lower bound of conditional MI between the output features of a positive pair.
\begin{equation}\label{eq:supcon_max_mi_samples}
   \log (2K-1) -\gL_{sup}(\vx,\vz;\mF,K) \leq \gI(\mF(\vx);\mF(\vz)|c)
\end{equation}
\end{proposition}


\begin{proposition}\label{prop:selfcon_loss}
SelfCon loss maximizes the lower bound of MI between the output features from the backbone and the sub-network.
\begin{equation}\label{eq:selfcon-s_max_mi_samples}
    \log (2K-1) - \gL_{self}(\vx;\{\mF,\mG\},K) \leq \gI(\mF(\vx);\mG(\vx)|c)
\end{equation}
\end{proposition}

SupCon and SelfCon loss have a negative sample size of $2(K-1)$ because of the augmented negative pairs for SupCon and the sub-network features for SelfCon. 

We extend the above MI bound to the MI between the intermediate and last feature of a backbone. Although MI is ill-defined between the variables with deterministic mapping, previous works view the training of a neural network as a stochastic process \citep{opening_black_box,goldfeld2019estimating,saxe2019information}. Thus, encoder features are considered as random variables, which allows us to define and analyze the MI between the features.

\begin{proposition}\label{prop:intermediate_feature}
As $\mF(\vx)$ and $\mG(\vx)$ are conditionally independent given the intermediate representation $\mT(\vx)$, they formulate a Markov chain: $\mG \leftrightarrow \mT \leftrightarrow \mF$ \citep{markov_chain}. Then, the following is satisfied.

\begin{equation}\label{eq:intermediate_feature}
    \gI(\mF(\vx);\mG(\vx)|c) \leq \gI(\mF(\vx);\mT(\vx)|c)
\end{equation}
\vspace{-8pt}
\end{proposition}

Proposition \ref{prop:intermediate_feature} states that minimizing SelfCon loss, which maximizes the lower bound of MI between the features from the backbone and the sub-network, can encourage the intermediate features to learn the class-related information from the last features.
Although there is indeed a gap in Eq. \ref{eq:intermediate_feature}, the gap between $\mathcal{I}(\bm{F}(\bm{x});\bm{G}(\bm{x})|c)$ and $\mathcal{I}(\bm{F}(\bm{x});\bm{T}(\bm{x})|c)$ may not be large since we implement $\bm{G}(\bm{x})$ as a simple linear transformation of $\bm{T}(\bm{x})$ in practice.
We empirically demonstrated the actual increment of $\mathcal{I}(\bm{F}(\bm{x});\bm{T}(\bm{x})|c)$ in Section \ref{exp:mi_estimation}.


\medskip
\paragraph{Discussion 4.2. How does increasing $\gI(\mF(\vx); \mT(\vx) |c)$ improve classification performance?}
To understand the information that SelfCon loss maximizes, we decompose the r.h.s. of Eq. \ref{eq:intermediate_feature} as follows:


\vspace{-10pt}
\begin{align}\label{eq:decomposed_mi}
    & \gI(\mF(\vx);\mT(\vx)|c) = \gI(\mF(\vx);\mT(\vx),c) - \gI(\mF(\vx);c) \\
    &= \underbrace{\gI(\mF(\vx);\mT(\vx))}_{(\square)} + \underbrace{\gI(\mF(\vx);c|\mT(\vx)) - \gI(\mF(\vx);c)}_{(\blacksquare)}. \nonumber
\end{align}

\noindent $(\square)$ implies that $\mT(\vx)$ is distilled with refined information (not conditional with respect to $c$) from $\mF(\vx)$, so the encoder can produce better representation \citep{infomax, local_dim}. On the other hand, $(\blacksquare)$ is interaction information \citep{infomeasure} that measures the influence of $\mT(\vx)$ on the amount of shared information between $\mF(\vx)$ and $c$. Increasing this interaction information means the intermediate feature enhances the correlation between the last feature and the label. 
Therefore, when we jointly optimize $(\square+\blacksquare)$, the intermediate and last features have aligned label information. 

In this sense, SelfCon loss is based on the \textit{InfoMax} principle \citep{linsker1989application}, which is about learning to maximize the MI between the input and output of a neural network. It has been proved that \textit{InfoMax}-based loss regularizes intermediate features and improves performance in semi-supervised \citep{ladder} and knowledge transfer \citep{vid} domains. Similar to the previous works, SelfCon loss increases the classification accuracy by regularizing the intermediate feature to have class-related information aligned with the last feature.

\medskip
\paragraph{Discussion 4.3. Is SelfCon loss applicable to unsupervised representation learning?}
The unsupervised version of SelfCon loss is a lower bound of $(\square)$ in Eq. \ref{eq:decomposed_mi}. By maximizing only $(\square)$, the last feature may follow the intermediate feature, learning redundant information about the input.\footnote{In supervision, a suboptimal case where $\mT(\vx)$ becomes a sink for $\mF(\vx)$ does not happen because the deeper layers have a larger capacity for label information \citep{opening_black_box}.} This could be the reason why SelfCon learning does \textit{not} work in an unsupervised environment (refer to Appendix \ref{app:self_sup}). However, to mitigate this problem, we propose in Appendix \ref{app:anchor_only_subnet} a loss function to prevent the backbone from following the sub-network. For this aim, we remove the term in Eq. \ref{eq:self-contrastive loss} where $\bm{\omega}=\mF$ (i.e., anchor from backbone) and $\bm{\omega}_j=\mG$ (i.e., contrastive pair from sub-network). This modification improves upon NT-Xent loss \citep{simclr} in the unsupervised CIFAR-100 experiment.

\section{Experiment}
\label{sec:experiment}

We present the image classification accuracy for standard benchmarks, such as CIFAR-10, CIFAR-100 \citep{cifar}, Tiny-ImageNet \citep{tiny-imagenet}, ImageNet-100 \citep{cmc}, and ImageNet \citep{imagenet}, and extensively analyze the results. We report the mean and standard deviation of top-1 accuracy over three random seeds. 
We used the optimal structure and position of the sub-network, however, the overall performance was comparable to or better than the baselines.
The complete implementation details and hyperparameter tuning results are presented in Appendix \ref{app:implementation details}. 

We also have implemented SupCon with a single-viewed batch (SupCon-S) and SelfCon with a multi-viewed batch (SelfCon-M) in order to examine the independent effects of the single-view and the sub-network. Note that their loss functions only require the change of the anchor set $I$ and corresponding positive and negative sets (i.e., $P_{ij}$ and $N_i$) in Eq. \ref{eq:supcon} and \ref{eq:self-contrastive loss}.


\subsection{Representation Learning}
\label{exp:representation}

We measured the classification accuracy of the representation learning protocol \citep{simclr}, which consists of \textit{2-stage training}: (1) pretraining an encoder network and (2) fine-tuning a linear classifier with the frozen encoder (called a linear evaluation).
In Appendix \ref{app:1-stage_training}, we compared with other supervised losses in the 1-stage training framework (i.e., not decoupling the encoder pretraining and fine-tuning).

\paragraph{Small-scale benchmark}
The classification accuracy is summarized in Table \ref{tab:represent_resnet}. Interestingly, the loss functions in the single-viewed batch outperform their multi-viewed counterparts in all settings. Furthermore, our SelfCon learning, which trains using the sub-network, shows higher classification accuracy than CE and SupCon. The effects of the sub-network are analyzed in Section \ref{exp:subnetwork}.

\begin{table*}[!t]
\small
\centering
\resizebox{0.97\textwidth}{!}{
\renewcommand{\arraystretch}{1.1}
\begin{tabular}{lcccccccc}
\toprule
                   &        &       & \multicolumn{3}{c}{ResNet-18}  & \multicolumn{3}{c}{ResNet-50} \\  \cmidrule(l{2pt}r{2pt}){4-6} \cmidrule(l{2pt}r{2pt}){7-9}
\multirow{-2.5}{*}{Method} & \multirow{-2.5}{*}{\thead{Single-\\View}} & \multirow{-2.5}{*}{\thead{Sub-\\Network}} &  CIFAR-10   &   CIFAR-100    &   Tiny-ImageNet  &  CIFAR-10   &   CIFAR-100    &   Tiny-ImageNet  \\ \midrule\hline
      CE      &      \checkmark    &  &   $94.7_{\pm{0.1}}$ & $72.9_{\pm{0.1}}$  & $57.5_{\pm{0.3}}$ & $94.9_{\pm{0.2}}$  &   $74.8_{\pm{0.1}}$    &   $62.3_{\pm{0.4}}$   \\ \hline
  SupCon    &           &     &    $94.7_{\pm{0.2}}$ & $73.0_{\pm{0.0}}$      & $56.9_{\pm{0.4}}$ & $95.6_{\pm{0.1}}^{\dag}$ & $75.5_{\pm{0.2}}^{\dag}$   & $61.6_{\pm{0.2}}$      \\
  SelfCon-M  &               & \checkmark &    $95.0_{\pm{0.1}}$ & $74.9_{\pm{0.1}}$      & $59.2_{\pm{0.0}}$  &   $95.5_{\pm{0.1}}$ & $76.9_{\pm{0.1}}$      & $63.0_{\pm{0.2}}$  \\
  SupCon-S\!   &        \checkmark     &    &   $94.9_{\pm{0.0}}$ & $73.9_{\pm{0.1}}$      & $58.4_{\pm{0.3}}$  &  $\mathbf{95.8_{\pm{0.1}}}$ & $76.7_{\pm{0.1}}$      & $62.0_{\pm{0.2}}$   \\
   \textbf{SelfCon}   &    \checkmark    & \checkmark  &   $\mathbf{95.3_{\pm{0.2}}}$ & $\mathbf{75.4_{\pm{0.1}}}$     & $\mathbf{59.8_{\pm{0.4}}}$  &   $\mathbf{95.7_{\pm{0.2}}}$ & $\mathbf{78.5_{\pm{0.3}}}$      & $\mathbf{63.7_{\pm{0.2}}}$  \\ \bottomrule
\end{tabular}}
\caption{The results of the linear evaluation for small-scale benchmarks. Bold type is for all the values of which the standard deviation range overlaps with that of the best accuracy. We used the same batch size of 1024 and a learning rate of 0.5 as \citet{supcon} did in CIFAR experiments. $\dag$: We have re-implemented SupCon and also run their official code for credibility, but the accuracy was slightly lower than their reported numbers.}\label{tab:represent_resnet}


\end{table*}

\begin{table*}[!t]
    \centering
    \small
    \addtolength{\tabcolsep}{-2.4pt}
    \resizebox{0.97\linewidth}{!}{
    \renewcommand{\arraystretch}{1.1}
    \begin{tabular}{lccccccccccccccccc}
    \toprule
    & & & \multicolumn{6}{c}{ImageNet-100} & \multicolumn{9}{c}{ImageNet} \\ \cmidrule(l{2pt}r{2pt}){4-9} \cmidrule(l{2pt}r{2pt}){10-18}
    & & &  \multicolumn{3}{c}{ResNet-18} & \multicolumn{3}{c}{ResNet-50} &  \multicolumn{3}{c}{ResNet-18} & \multicolumn{3}{c}{ResNet-34} & \multicolumn{3}{c}{ResNet-50} \\ \cmidrule(l{2pt}r{2pt}){4-6} \cmidrule(l{2pt}r{2pt}){7-9} \cmidrule(l{2pt}r{2pt}){10-12} \cmidrule(l{2pt}r{2pt}){13-15} \cmidrule(l{2pt}r{2pt}){16-18}
    \multirow{-4}{*}{Method} & \multirow{-4}{*}{\thead{Single-\\View}} & \multirow{-4}{*}{\thead{Sub-\\Network}} & Mem. & Time & Acc. & Mem. & Time & Acc. & Mem. & Time & Acc. & Mem. & Time & Acc. & Mem. & Time & Acc. \\ \midrule \hline
    CE & \checkmark & & - & - & $83.7$ & - & - & $86.4$ & - & - & $69.4$ & - & - & $72.7$ & - & - & $76.5^{\ddag}$ \\  
    \hline
    SupCon & & &  $\times 1.5$ & $\times 2.1$ & $85.6$ & $\times 1.7$ & $\times 1.9$ & $88.2$ & $\times 1.5$ & $\times 2.2$ & $71.2$ & $\times 1.5$ & $\times 2.1$ & $74.9$ & $\times 1.7$ & $\times 2.1$ & $78.0^{\ddag}$  \\
    SelfCon-M  & & \checkmark &  $\times 1.6$ & $\times 2.1$ & $85.8$ & $\times 1.8$ & $\times 2.2$ & $\mathbf{88.7}$ & $\times 1.7$ & $\times 2.3$ & $\mathbf{71.6}$ & $\times 1.7$ & $\times 2.2$ & $75.5$  & $\times 1.8$ & $\times 2.2$ & $78.4$ \\
    SupCon-S  & \checkmark & &  $\times \mathbf{1.0}$ & $\times \mathbf{1.0}$ & $84.9$ & $\times \mathbf{0.9}$ & $\times \mathbf{0.8}$ & $87.8$ & $\times \mathbf{0.9}$ & $\times \mathbf{1.0}$ & $70.2$ & $\times \mathbf{0.9}$ & $\times \mathbf{1.0}$ & $74.4$ & $\mathbf{\times 0.9}$ & $\mathbf{\times 0.9}$ & $77.5$ \\
    \textbf{SelfCon}  & \checkmark & \checkmark &  \underline{$\times \mathbf{1.0}$} & \underline{$\times \mathbf{1.0}$} & $\mathbf{86.1}$ & \underline{$\times 1.0$} & \underline{$\times 1.0$} & $\mathbf{88.7}$ & \underline{$\times 1.0$} & \underline{$\times \mathbf{1.0}$} & $71.4$ & \underline{$\times 1.0$} & \underline{$\times \mathbf{1.0}$} & $\mathbf{75.6}$ & \underline{$\times 1.0$} & \underline{$\times 1.0$} & $\mathbf{78.6}$ \\
    \bottomrule
    \end{tabular}}
    \caption{The classification accuracy for ImageNet-100 and ImageNet. 
    We summarized the ratio of memory (GiB / GPU) and time (sec / step) based on those of SelfCon in each architecture. $\ddag$: We used the results in the same setting as ours (e.g., $B=1024$) reported by \citet{supcon} (refer to Figure 4 in their original paper).}\label{tab:imagenet_total}
    \vspace{-5pt}
\end{table*}

\paragraph{Large-scale benchmark}
We summarized the experimental results for the ImageNet-100, of which 100 classes were randomly sampled \citep{cmc}, and the full-scale ImageNet (Table \ref{tab:imagenet_total}). 
Our SelfCon learning that includes the sub-network consistently outperforms SupCon learning on both ImageNet-100 and ImageNet. In particular, SelfCon showed a higher efficiency ratio (i.e., cost-to-accuracy) than SupCon, SelfCon-M, and SupCon-S.
Different from small-scale benchmarks, we observed that the training difficulty of large-scale images could degrade the performance of the single-viewed method (see SupCon-S vs. SupCon). 
The poor performance of SupCon-S, which consumes an amount of memory and time similar to SelfCon, reflects the superiority of SelfCon.

On large-scale benchmarks, the difference in accuracy between SelfCon and SelfCon-M was smaller than that on small-scale benchmarks.
We suppose that it is mainly attributed to the over-/under-fitting problem.
In fact, various factors (e.g., architecture, dataset, batch size, and training epochs) in combination can affect the bias-variance trade-off. 
For example, the ImageNet result on ResNet-18 appears to be affected by the underfitting from a small architecture and a huge dataset (also refer to Appendix \ref{app:imagenet_small_batch}). 
We intensively analyzed the effects of different factors in terms of the single-view and multi-view in Section \ref{exp:single_view}.

\paragraph{Ensemble prediction with sub-network}

The co-trained sub-network is a novel strength of SelfCon learning as an efficient and simple boosting technique.
In practice, training an extra linear classifier after the frozen sub-network does not demand a high cost in the fine-tuning scheme. We can thus obtain two additional linear evaluation results by (1) fine-tuning a classifier after the sub-network output and (2) ensembling the predictions of two classifiers. Table \ref{tab:subnet_ensemble} indicates that the ensemble prediction is the most powerful technique we have proposed.
In particular, SelfCon can achieve a significant performance gain of +3.0\% on ImageNet without requiring cost-intensive techniques such as multi-viewed batch or larger batch size \citep{simclr, supcon}. Refer to Appendix \ref{app:ensemble_resnet50} for the results on ResNet-18.

\begin{table}[!t]
\centering
\scriptsize
\vspace{5pt}
\addtolength{\tabcolsep}{-2pt}
\resizebox{.875\linewidth}{!}{
\renewcommand{\arraystretch}{0.8}
\begin{tabular}{lcccc}
\toprule
Method & CF-100 & Tiny-IN & IN-100 & IN \\ \midrule \hline
CE & $74.8$ & $62.3$ & $86.4$ & $76.5$ \\ 
SupCon & $75.5$ & $61.6$ & $88.2$ & $78.0$ \\ \hline 
Backbone & $78.5$ & $63.7$ & $88.7$ & $78.6$ \\
Sub-network  & $73.3$ & $58.9$ & $87.6$ & $78.5$ \\
\textbf{Ensemble}  & $\mathbf{80.0}$ & $\mathbf{65.7}$ & $\mathbf{89.1}$ & $\mathbf{79.5}$ \\
${\,}^\llcorner$Gain (vs. CE)  & +$5.2$ & +$3.4$ & +$2.7$ & +$3.0$ \\
${\,}^\llcorner$Gain (vs. SupCon) & +$4.5$ & +$4.1$ & +$0.9$ & +$1.5$ \\
\bottomrule
\end{tabular}}
\caption{Classification accuracy with the classifiers after backbone, sub-network, and the ensemble of them. The ResNet-50 encoder is pretrained by the SelfCon loss function.}
\vspace{-15pt}
\label{tab:subnet_ensemble}
\end{table}

\paragraph{Downstream tasks}
Thus far, we have observed the SelfCon's superiority via linear evaluation performance. While our main goal is supervised classification on the target dataset, we can further use the pretrained encoder to transfer to other downstream tasks. Hence, in Table \ref{tab:downstream_tasks}, we summarized the results of the downstream tasks, eight fine-grained recognition datasets and two semantic segmentation or object detection datasets, to further verify the transferability of the SelfCon's pretrained encoder.
SelfCon outperforms SupCon in most of the downstream tasks, implying that ImageNet-pretrained SelfCon contains more generalized representation. Specifically, SelfCon greatly improves up to +6.8\% and +4.4\% for fine-grained and semantic segmentation tasks, respectively. 

\begin{table*}[!t]
    \begin{subfigure}[b]{0.5\linewidth}
    \centering
    \scriptsize
    \addtolength{\tabcolsep}{-3.5pt}
    \resizebox{0.93\linewidth}{!}{
    \renewcommand{\arraystretch}{0.8}
    \begin{tabular}{lcccccccc}
    \toprule
        Method & CUB & Dogs & MIT67 & Flowers & Pets & \!\!Stanford40\!\! & Cars & Aircraft \\ \midrule \hline
        \textbf{SelfCon} & $\mathbf{62.2}$ & $91.8$ & $\mathbf{72.3}$ & $\mathbf{85.7}$ & $\mathbf{90.6}$ & $\mathbf{77.5}$ & $\mathbf{45.2}$ & $\mathbf{39.0}$ \\
        SupCon & $56.8$ & $\mathbf{92.3}$ & $65.5$ & $82.8$ & $89.6$ & $76.7$ & $40.4$ & $37.6$ \\
        \bottomrule
    \end{tabular}}
    \caption{Fine-grained Recognition}
    \end{subfigure}%
    \begin{subfigure}[b]{0.24\linewidth}
    \centering
    \scriptsize
    \addtolength{\tabcolsep}{-3pt}
    \resizebox{0.93\linewidth}{!}{
    \renewcommand{\arraystretch}{0.8}
    \begin{tabular}{lcc}
    \toprule
        Method & Pascal VOC & MS COCO  \\ \midrule \hline
        \textbf{SelfCon} & $\mathbf{71.6}$ &  $\mathbf{48.1}$ \\
        SupCon & $69.6$ & $43.7$ \\
        \bottomrule
    \end{tabular}}
    \caption{Semantic Segmentation}
    \end{subfigure}
    %
    \begin{subfigure}[b]{0.24\linewidth}
    \centering
    \scriptsize
    \addtolength{\tabcolsep}{-3pt}
    \resizebox{0.93\linewidth}{!}{
    \renewcommand{\arraystretch}{0.8}
    \begin{tabular}{lcc}
    \toprule
        Method & Pascal VOC & MS COCO  \\ \midrule \hline
        \textbf{SelfCon} & $\mathbf{63.0}$ &  $\mathbf{29.4}$ \\
        SupCon & $61.8$ & $28.8$ \\
        \bottomrule
    \end{tabular}}
    \caption{Object Detection}
    \end{subfigure}
    \vspace{-5pt}
    \caption{Downstream task results of SelfCon and SupCon encoders. The ResNet-34 model pretrained on ImageNet is transferred. The evaluation metric is (a) linear evaluation accuracy, (b) mIoU, and (c) mAP. For the semantic segmentation, we used a DeepLabV3+ module \citep{deeplabv3+}, and for the object detection, we used a RetinaNet detector \citep{retinanet}. The dataset details are in Appendix \ref{app:implementation details}.}
    \label{tab:downstream_tasks}
\end{table*}

\begin{figure*}[!t]
\vspace{-5pt}
    \begin{subfigure}[b]{0.253\linewidth}
    \centering
    \includegraphics[width=\linewidth]{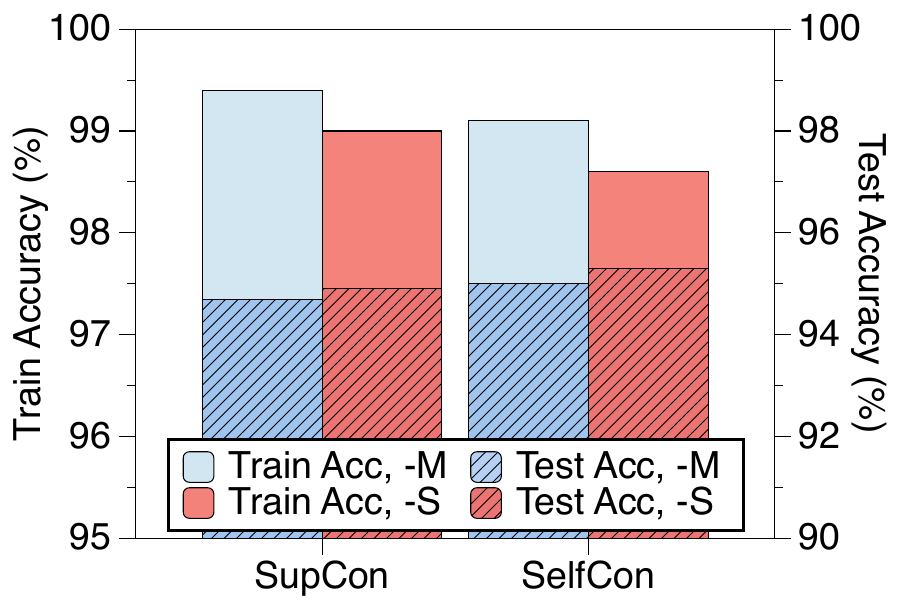}
    \caption{\small {CIFAR-10}}\label{fig:generalization_error_a}
    \end{subfigure}%
    \hfill
    \begin{subfigure}[b]{0.253\linewidth}
    \centering
    \includegraphics[width=\linewidth]{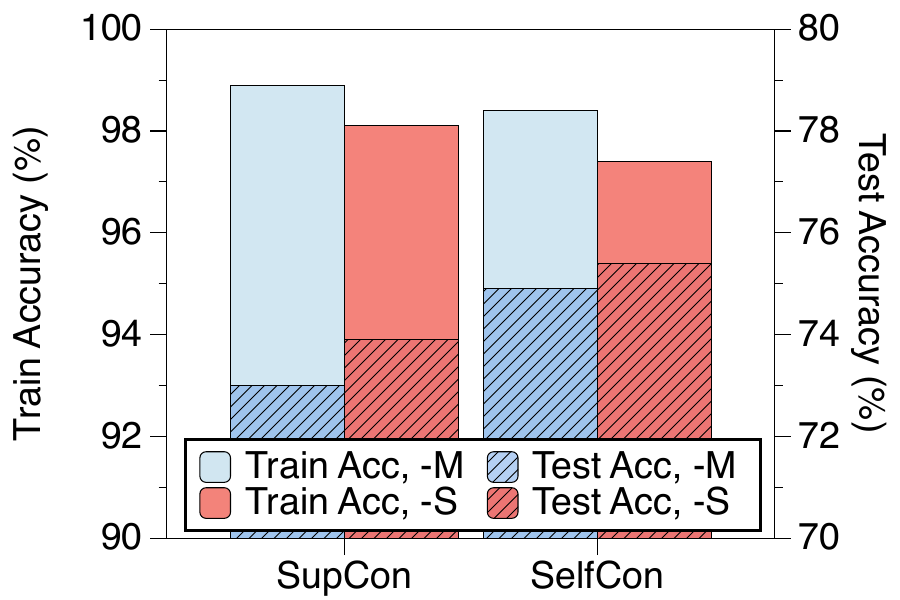}
    \caption{\small {CIFAR-100}}\label{fig:generalization_error_b}
    \end{subfigure}
    \hfill
    \begin{subfigure}[b]{0.253\linewidth}
    \centering
    \includegraphics[width=\linewidth]{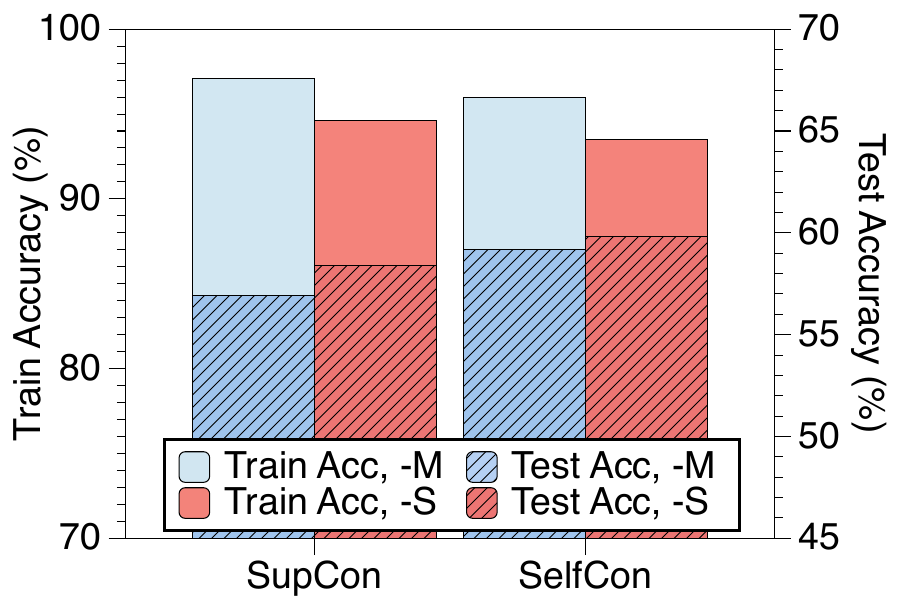}
    \caption{\small {Tiny-ImageNet}}\label{fig:generalization_error_c}
    \end{subfigure}
    \hfill
    \begin{subfigure}[b]{0.225\linewidth}
    \centering
    \includegraphics[width=\linewidth]{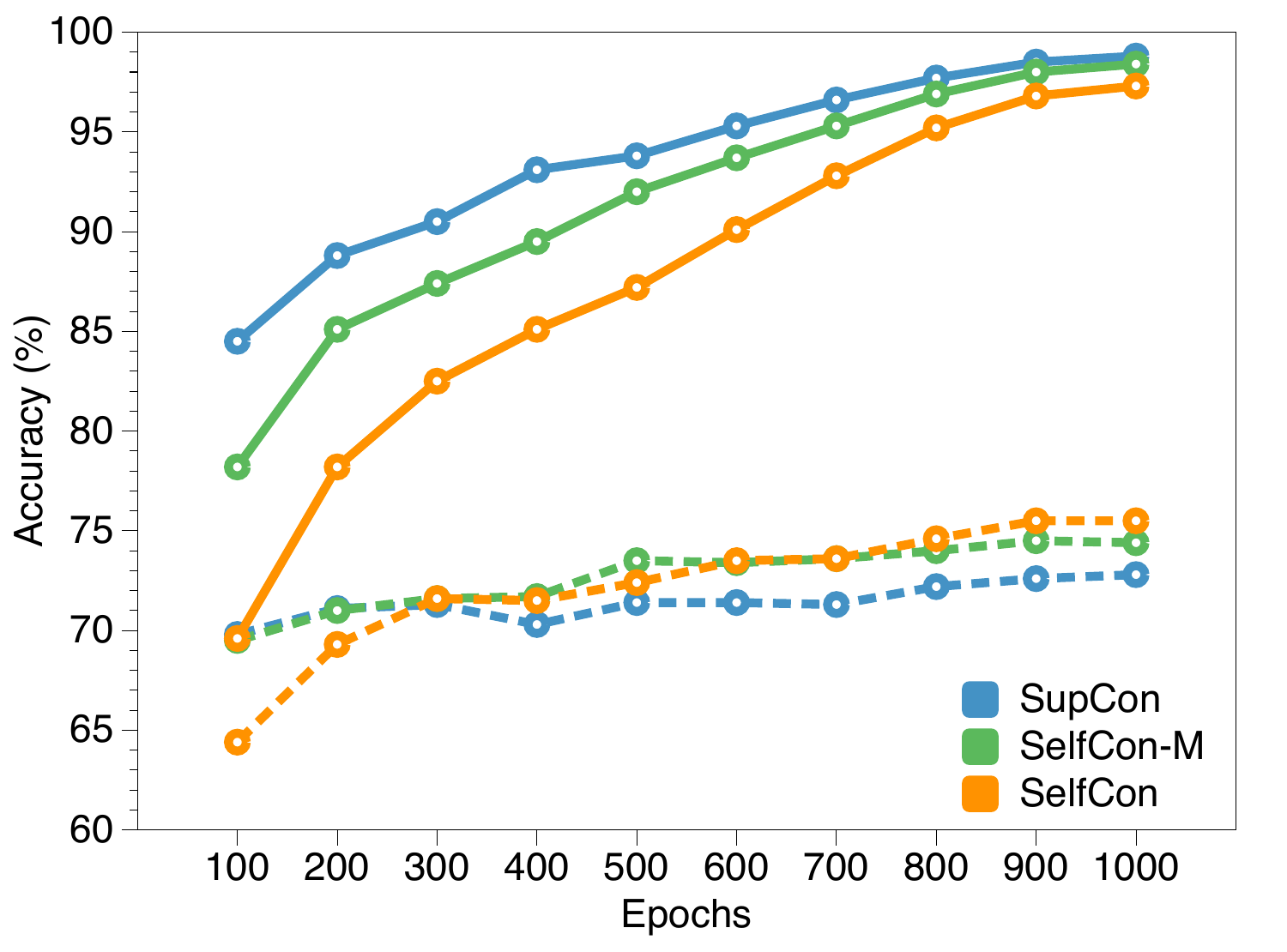}
    \caption{\small {Training Process}}\label{fig:memorization_test}
  \end{subfigure}
\vspace{-5pt}
\caption{(a--c) The train and test accuracy on ResNet-18 for different views and loss functions. The accuracy is measured with a linear classifier during the linear evaluation. (d) CIFAR-100 accuracy on ResNet-18 at different epochs. The solid and dashed lines are for train and test accuracy, respectively.}
\label{fig:generalization_error}
\vspace{-5pt}
\end{figure*}

\vspace{-2pt}
\subsection{Single-view vs. Multi-view}
\label{exp:single_view}

\paragraph{Single-view reduces generalization error.}
In Figures \ref{fig:generalization_error_a}-\ref{fig:generalization_error_c}, SupCon shows higher train accuracy, but lower test accuracy than SupCon-S, and the same trend is observed with SelfCon-M and SelfCon (blue vs. red). Compared with single-view, multi-view from the augmented image makes the encoder amplify the memorization of data and results in overfitting to each instance. In addition, Figure \ref{fig:memorization_test} shows that SelfCon gradually enhances generalization ability, while SelfCon-M and SupCon achieve a little gain in test accuracy despite the fast convergence.

\paragraph{Multi-view is advantageous for small batch size.}


In supervised learning, a large batch size has been known to reduce generalization ability, which degrades performance \citep{large_batch_cnn, large_batch_bn, large_batch_sgd}. We examined whether the performance in a supervised contrastive framework is also dependent on the batch size. In Table \ref{tab:smaller_batch}, SelfCon showed the best performance in every case except for the batch size of 64. However, the multi-viewed method outperformed the single-viewed counterpart in 64-batch experiments, where underfitting may occur because of large randomness from the small batch size or the small number of positive pairs. 
In the ImageNet experiment on ResNet-18 (see Table \ref{tab:imagenet_total}), SelfCon-M also outperformed every method, implying that it is more important to mitigate underfitting for large-scale dataset. Conversely, in ResNet-34 and ResNet-50, SelfCon showed the best performance.
In summary, multi-viewed methods may have good performance in the underfitting scenario (e.g., small batch size, small epochs, or large-scale benchmark).

\paragraph{Single-view is efficient in terms of memory usage and computational cost.}
To investigate the efficiency of a single-viewed batch against a conventional multi-viewed batch, we have compared the required memory and time cost in Table \ref{tab:imagenet_total}. 
Due to the additional augmented samples, the computational cost of the multi-viewed approaches is around twice as much as their single-viewed counterparts. SelfCon, on the other hand, outperformed every method with a low cost under a range of experimental conditions, while SupCon-S showed poor performance in the large-scale benchmarks.
In Appendix \ref{app:memory}, we summarized the detailed numbers of the costs for SelfCon and SupCon. Although SelfCon requires the additional parameters owing to the sub-network, its memory and computation cost in practice are much more efficient. 

\begin{table}[tb]
\centering
\scriptsize
\vspace{5pt}
\addtolength{\tabcolsep}{-0pt}
\resizebox{0.875\linewidth}{!}{
\renewcommand{\arraystretch}{0.8}
    \begin{tabular}{lccccc}
    \toprule
    & \multicolumn{5}{c}{Batch Size} \\
    \cmidrule(l{2pt}r{2pt}){2-6}
    \multirow{-2.5}{*}{Method} & $64$ & $128$ & $256$ & $512$ & $1024$ \\ \midrule \hline
    CE & $74.9$ & $74.9$ & $74.1$ & $73.3$ & $72.9$ \\
    \hline
    SupCon & $74.8$ & $73.8$ & $72.9$ & $72.5$ & $73.0$  \\
    SelfCon-M & $\mathbf{75.8}$ & $76.5$ & $75.9$ & $75.0$ & $74.9$  \\
    SupCon-S & $73.6$& $75.3$ & $75.0$ & $74.0$ & $73.9$  \\
    SelfCon &  $74.0$ & $\mathbf{76.6}$ & $\mathbf{77.0}$ & $\mathbf{75.8}$ & $\mathbf{75.4}$ \\ \bottomrule
    \end{tabular}}
\caption{The classification accuracy of CIFAR-100 on ResNet-18 with various batch sizes.}
\label{tab:smaller_batch}
\vspace{-10pt}
\end{table}

\subsection{What Does the Sub-network Achieve?}
\label{exp:subnetwork}




\paragraph{Regularization effect}

SelfCon loss regularizes the sub-network to output similar features to the backbone network. It prevents the encoder from overfitting to the data, and it is effective in multi-viewed as well as single-viewed batches. In Figures \ref{fig:generalization_error_a}--\ref{fig:generalization_error_c}, we confirm the regularization effect (i.e., lower train accuracy, but higher test accuracy) by comparing each bar of the same color. The strong regularization of the sub-network helped SelfCon (also with multi-view) outperform the SupCon counterparts. This trend can also be observed in Figure \ref{fig:memorization_test} and Table \ref{tab:smaller_batch}.




\begin{figure}[!t]
\centering
\vspace{-5pt}
\includegraphics[width=\linewidth]{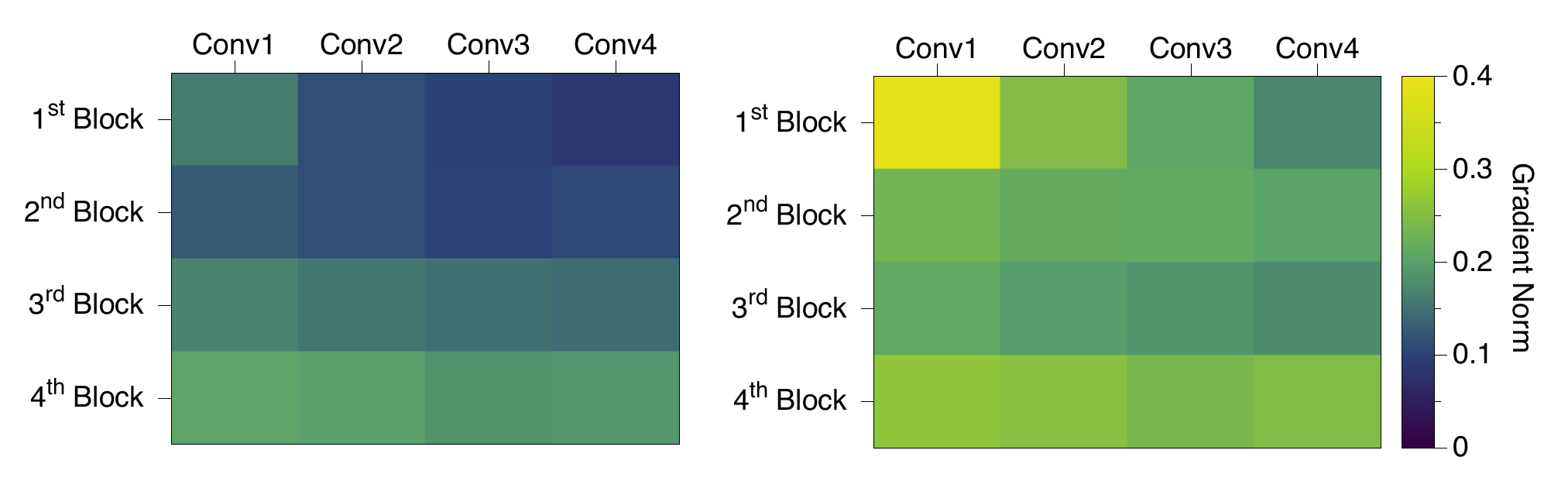}
\vspace{-15pt}
\caption{Gradient norm of each ResNet-18 block and convolutional layer. We computed gradients from the SupCon loss (Left) and SelfCon-M loss (Right), both from the same initialized model. All convolution layers in the block are named by order.}
\vspace{-5pt}
\label{fig:vanishing_grad}
\end{figure}

\begin{figure}[!t]
    \centering
    \includegraphics[width=\linewidth]{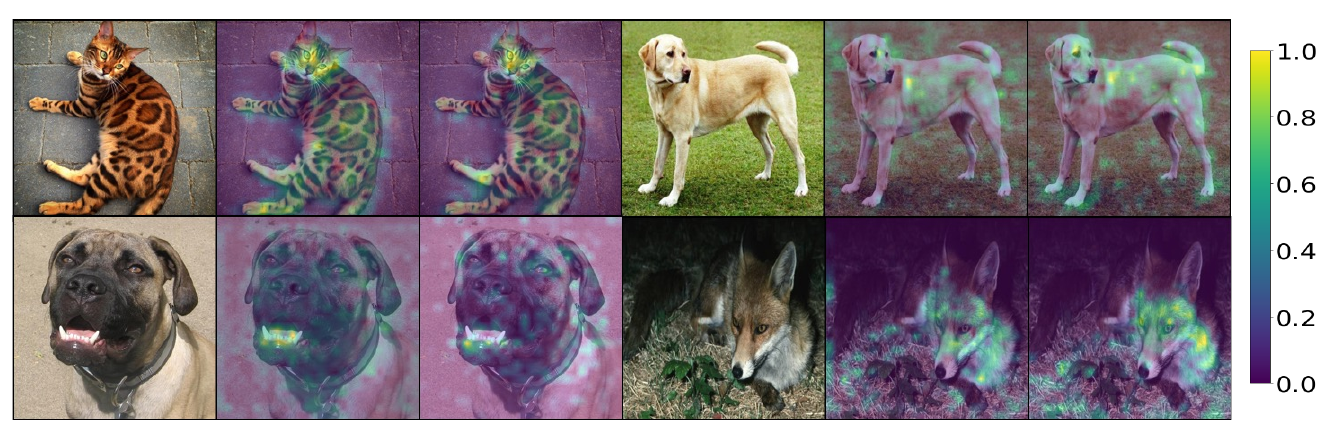}
    \vspace{-15pt}
    \caption{Grad-CAM \citep{gradcam} visualizations for the feature-level multi-view generated by the sub-network. Along with the original image, each map visualizes the gradients from the sub-network (Left) and the backbone network (Right), respectively.}
    \label{fig:feature_multiview}
\end{figure}



\paragraph{Mitigating the vanishing gradient}
SelfCon learning sends more abundant information to the earlier layers through the gradients flowing from the sub-networks.
Previous works \citep{deeply_supervised,branchynet,be_your_own} also have pointed out that the success of the multi-exit framework owes to solving the vanishing gradient problem.
In Figure \ref{fig:vanishing_grad}, a large gradient flows up to the earlier layer in the SelfCon-M, whereas a large amount of the SupCon loss gradient vanishes. 
Note that the sub-network is positioned after the $2^{\text{nd}}$ block of the ResNet-18 backbone network. 
Thus, there is a significant difference in the gradient norm in the $2^{\text{nd}}$ block of the encoder. 

\paragraph{Feature-level multi-view}

One of the advantages of SelfCon learning is that it relaxes the dependency on multi-viewed batches. This is accomplished by the multi-views on the representation space made by the parameters of the sub-network.
In Figure \ref{fig:feature_multiview}, we visualize the gradient of SelfCon loss w.r.t. the intermediate layer of the backbone network (ResNet-18), right before the exit path. Both networks focus on similar but clearly different pixels of the same input image, implying that the sub-network learns another view in the feature space.
As the multi-view in contrastive learning requires domain-specific augmentation, recent studies have explored domain-agnostic methods of augmentation \citep{i-mix,dacl}. SelfCon could be an intriguing future work in that auxiliary networks could be an efficient substitute for data augmentation.

\subsection{Mutual Information Estimation}
\label{exp:mi_estimation}




\begin{figure}
\centering
\includegraphics[width=.9\linewidth]{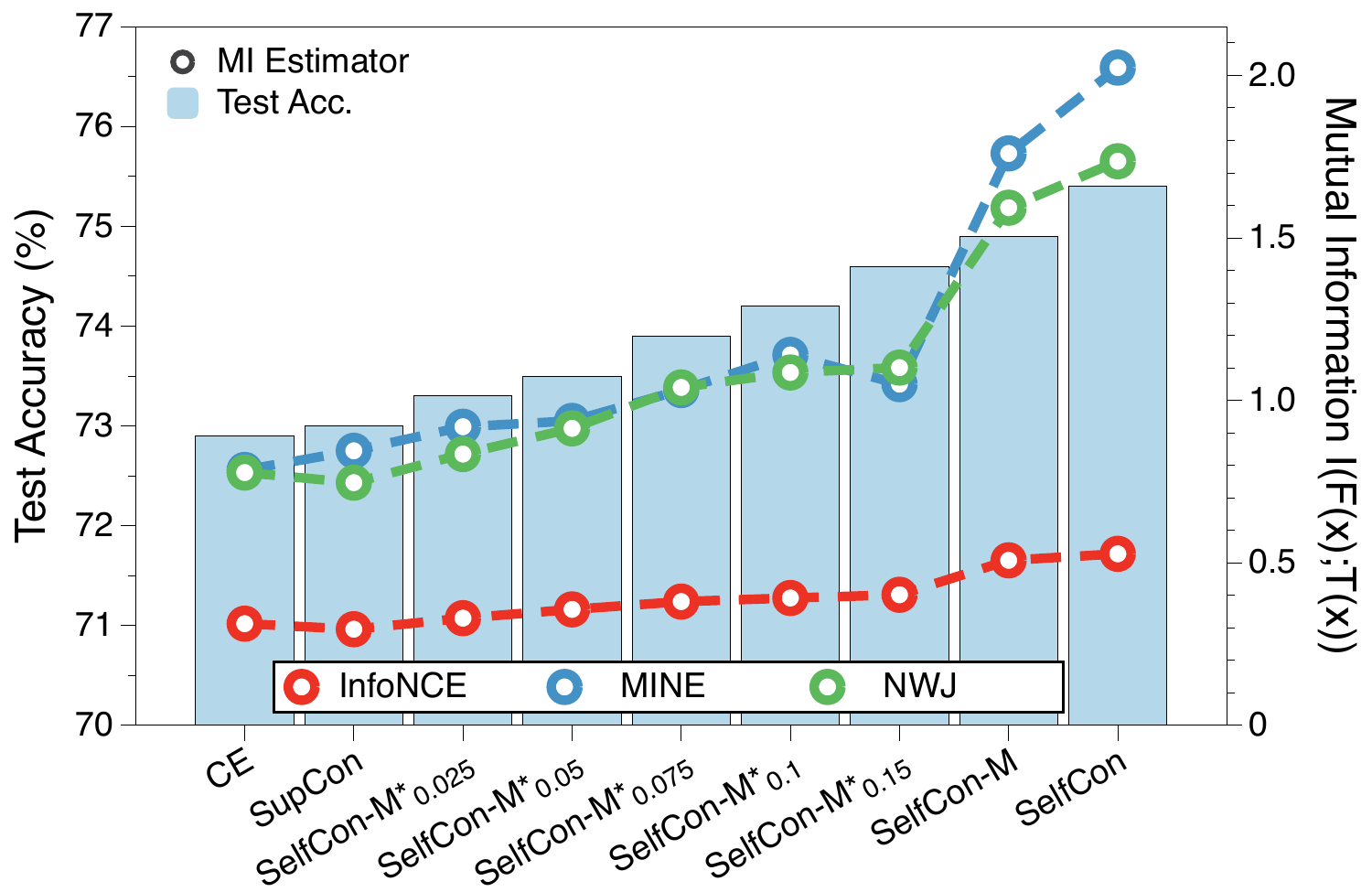}
\vspace{-5pt}
\caption{Test accuracy and the estimated mutual information of different methods. SelfCon-M*$_\alpha$ denotes SelfCon-M* loss with hyperparameter $\alpha$. 
When $\alpha \geq 0.2$, the test accuracy was similar to that of SelfCon-M.}
\label{fig:mi_connect_correlation}
\end{figure}

We argue that minimizing SelfCon loss maximizes the lower bound of MI, which results in the improved classification performance presented in Section \ref{sec:discussions}. To empirically confirm this claim, we design an interpolation between SupCon and SelfCon-M loss as follows:
\begin{equation}
    \mathcal{L}_{self\text{-}m*} = \frac{1}{1 + \alpha}\,\mathcal{L}_{sup} + \frac{\alpha}{1 + \alpha}\,\mathcal{L}_{self\text{-}m}\Big|_{\bm{\omega}=\mG}
\end{equation}
If $\alpha = 0$, $\mathcal{L}_{self\text{-}m*}$ is equivalent to the SupCon loss, and if $\alpha = 1$, then $\mathcal{L}_{self\text{-}m*}$ is \textit{almost} the same as SelfCon-M loss. We cannot make the exact interpolation because SelfCon-M has contrastive pairs from the sub-network, whereas SupCon does not.


In Figure \ref{fig:mi_connect_correlation}, we estimated MI with ResNet-18 and CIFAR-100 using various estimators: InfoNCE \citep{cpc}, MINE \citep{mine}, and NWJ \citep{nwj}. We measured $\gI(\mF(\vx);\mT(\vx))$ because it is difficult to estimate the conditional MI. We observed a clear increasing trend for both MI and the test accuracy as the contribution of SelfCon becomes larger (i.e., increasing $\alpha$). After SelfCon loss increases the correlation between $\mF(\vx)$ and $\mT(\vx)$, the rich information in earlier features enables the encoder to output a better representation because the intermediate feature is also the input for the subsequent layers. Refer to Appendix \ref{app:mi_details} for a detailed SelfCon-M* loss formulation and the exact numbers.

\section{Conclusion}
\label{sec:conclusion}

We have proposed a single-viewed supervised contrastive framework called Self-Contrastive learning, which self-contrasts the multiple features from a multi-exit architecture.
By replacing the augmentation with the sub-network, SelfCon enables the encoder to contrast within multiple features from a single image while significantly reducing the computational cost.
We verified by extensive experiments that SelfCon loss outperforms CE and SupCon loss.
We analyzed the success of SelfCon learning by exploring the effect of single-view and sub-network, such as the regularization effect, computational efficiency, or ensemble prediction.
In addition, we theoretically proved that SelfCon loss regularizes the intermediate features to learn the label information in the last feature, as our MI estimation experiment has supported.

\clearpage
\section*{Acknowledgments}
This research was supported by SK Hynix AICC K20.06\_Unsupervised DL Model Error Estimation\,(90\%). This work was also supported by Institute of Information \& communications Technology Planning \& Evaluation (IITP) grant funded by the Korea government(MSIT) (No.2019-0-00075, Artificial Intelligence Graduate School Program(KAIST), 10\%).

\bibliography{aaai23}

\clearpage
\appendix
\onecolumn
\clearpage


\section{Proofs}
\label{app:proofs}

\subsection{Proof of Proposition \ref{prop:supcon_loss}}
\begin{proof}
We extend the exact bound of InfoNCE \citep{interpolation_nce_nwj,demi}. Here, we consider the supervised setting where there are $C$ training classes. Without loss of generality, choose a class $c$ out of $C$ classes, and let $\vx$ and $\vz$ be different samples that share the same class label $c$. The derivation for the multi-view ($\vz$ being an augmented sample of $\vx$) is similar. For conciseness of the proof, we consider that no other image in a batch shares the same class. We prove that minimizing the SupCon loss \citep{supcon} maximizes the lower bound of conditional MI between two samples $\vx$ and $\vz$ given the label $c$:

\begin{equation}
    \mathcal{I}(\vx;\vz|c) \geq \log (2K-1) -\mathcal{L}_{sup}(\vx,\vz;\mF,K)
\end{equation}

\noindent for some function $\mF$ and hyperparameter $K$.

We start from Barber and Agakov's variational lower bound on MI \citep{variational_mi}.

\begin{equation}
    \mathcal{I}(\vx;\vz|c) = \mathbb{E}_{p(\vx,\vz|c)} \log \frac{p(\vz|\vx,c)}{p(\vz|c)} \geq \mathbb{E}_{p(\vx,\vz|c)} \log \frac{q(\vz|\vx,c)}{p(\vz|c)}
\end{equation}

\noindent where $q$ is a variational distribution. Since $q$ is arbitrary, we can set the sampling strategy as follows. First, sample $\vz_1$ from the proposal distribution $\pi(\vz|c)$ where $c$ is a class label of $\vx$. Then, sample $(K-1)$ negative samples $\{\vz_2,\cdots,\vz_K\}$ from the distribution $\sum_{c' \neq c} \pi(\vz,c')$, so that these negative samples do not share the class label with $\vx$. We augment each negative sample by random augmentation and concatenate with the original samples, i.e., $\{\vz_2, \cdots, \vz_K, \vz_{K+1},\cdots,\vz_{2K-1}\}$, where $\vz_{K+i-1}$ is the augmented sample from $\vz_i$ for $2\leq i \leq K$. We define the unnormalized density of $\vz_1$ given a specific set $\{\vz_2,\cdots,\vz_{2K-1}\}$ and $\vx$ of label $c$ is

\begin{equation}\label{eq:placeholder10}
    q(\vz_1|\vx,\vz_{2:(2K-1)},c) = \pi(\vz_1|c) \cdot \frac{(2K-1)\cdot e^{\psi_\mF(\vx,\vz_1)}}{e^{\psi_\mF(\vx,\vz_1)} + \sum_{k=2}^{2K-1} e^{\psi_\mF(\vx,\vz_k)}}
\end{equation}

\noindent where $\psi$ is often called a discriminator function \citep{infomax}, defined as $\psi_\mF(\vu,\vv)=\mF(\vu) \cdot \mF(\vv)$ for some vectors $\vu, \vv$. By setting the proposal distribution as $\pi(\vz|c)=p(\vz|c)$, we obtain the MI bound:
\begin{align}
    \mathcal{I}(\vx;\vz|c) \label{eq:placeholder1}
    & \geq \mathbb{E}_{p(\vx,\vz_1|c)} \log \frac{q(\vz_1|\vx,c)}{p(\vz_1|c)} \\
    & = \mathbb{E}_{p(\vx,\vz_1|c)} \log \dfrac{\mathbb{E}_{p(\vz_{2:(2K-1)}|c)} q(\vz_1|\vx,\vz_{2:(2K-1)},c)}{p(\vz_1|c)} \\
    & \geq \mathbb{E}_{p(\vx,\vz_1|c)} \Bigg[ \mathbb{E}_{p(\vz_{2:(2K-1)}|c)} \log \frac{ p(\vz_1|c) \cdot \frac{(2K-1) \cdot e^{\psi_\mF(\vx,\vz_1)}}{e^{\psi_\mF(\vx,\vz_1)} + \sum_{k=2}^{2K-1} e^{\psi_\mF(\vx,\vz_k)}}}{p(\vz_1|c)} \Bigg] \\
    & = \mathbb{E}_{p(\vx,\vz_1|c) p(\vz_{2:(2K-1)}|c)} \log \frac{ e^{\psi_\mF(\vx,\vz_1)}}{\frac{1}{2K-1} \sum_{k=1}^{2K-1} e^{\psi_\mF(\vx,\vz_k)}} \label{eq:placeholder6}\\
    & = \log (2K-1) - \mathcal{L}_{sup}(\vx,\vz;\mF,K).\label{eq:placeholder2}
\end{align}

\noindent where the second inequality is derived from Jensen's inequality. Because Eq. \ref{eq:placeholder6} is an expectation with respect to the sampled $\vx$ and $\vz_1$, the case where the anchor is swapped to $\vz_1$ is also being considered.

A neural network $\mF$ (backbone in our framework) with $L$ layers are formulated as $\mF = f_L \circ f_{L-1} \circ \cdots \circ f_1$. Then, $\psi_\mF(\vu,\vv) = \mF(\vu) \cdot \mF(\vv) = f_{1:L}(\vu) \cdot f_{1:L}(\vv)$. We define another discriminator function as $\psi^{\dag}_\mF (\vu,\vv) = f_{(\ell+1):L}(\vu) \cdot f_{(\ell+1):L}(\vv)$. Obviously, the following equivalence holds:

\begin{equation}
    \psi^{\dag}_\mF (f_{1:\ell}(\vu), f_{1:\ell}(\vv)) = \psi_\mF(\vu, \vv).
\end{equation}

Note that $f_{1:\ell}(\vu)$ is the $\ell$-th intermediate feature of input $\vu$.
Following the same procedure as in Eq. \ref{eq:placeholder1}-\ref{eq:placeholder2},
\begin{align}
    \mathcal{I}(f_{1:\ell}(\vx); f_{1:\ell}(\vz) |c)
    & \geq \mathbb{E}_{p(\vx,\vz_1|c) p(\vz_{2:(2K-1)}|c)} \log \frac{e^{\psi^{\dag}_\mF (f_{1:\ell}(\vx), f_{1:\ell}(\vz_1))}}{\frac{1}{2K-1} \sum_{k=1}^{2K-1} e^{\psi^{\dag}_\mF (f_{1:\ell}(\vx), f_{1:\ell}(\vz_k))}} \\
    & = \mathbb{E}_{p(\vx,\vz_1|c) p(\vz_{2:(2K-1)}|c)} \log \frac{e^{\psi_\mF(\vx,\vz_1)}}{\frac{1}{2K-1} \sum_{k=1}^{2K-1} e^{\psi_\mF(\vx,\vz_k)}} \\
    & = \log (2K-1) - \mathcal{L}_{sup}(\vx,\vz;\mF,K)
\end{align}
\vspace{-10pt}

From above, as the intermediate feature is arbitrary to the position, we can obtain a similar inequality:

\begin{align}
    \mathcal{I}(f_{(\ell+1):L}(f_{1:\ell}(\vx)); f_{(\ell+1):L}(f_{1:\ell}(\vz)) |c)
    & = \mathcal{I}(\mF(\vx); \mF(\vz) |c) \\
    & = \mathcal{I}(f_{1:L}(\vx); f_{1:L}(\vz) |c) \\
    & \geq \log (2K-1) -\mathcal{L}_{sup}(\vx,\vz; \mF,K).
\end{align}

\end{proof}

\subsection{Proof of Proposition \ref{prop:selfcon_loss}}
\label{app:prop_selfcon_loss}
\begin{proof}

In Proposition \ref{prop:supcon_loss}, we proved that SupCon loss maximizes the lower bound of conditional MI between the output features of a positive pair. We can think of another scenario where the network $\mF$ now has a sub-network $\mG$. Assume that the sub-network has $M>\ell$ layers: $\mG = g_M \circ g_{M-1} \circ \cdots \circ g_1$. As we discussed in the paper, the exit path is placed after the $\ell$-th layer, so regarding our definition of the sub-network, $\mG$ shares the same parameters with $\mF$ up to $\ell$-th layer, i.e., $g_1=f_1$, $g_2=f_2$, $\cdots$, $g_{\ell}=f_{\ell}$. Define $\psi_\mG(\vu,\vv) = \mG(\vu) \cdot \mG(\vv)$

We introduce a discriminator function that measures the similarity between the outputs from the backbone and the sub-network, $\psi_{\mF\mG}(\vu,\vv)=\mF(\vu) \cdot \mG(\vv)$. Similarly, $\psi_{\mG\mF}(\vu,\vv)=\mG(\vu) \cdot \mF(\vv)$. Considering that the SelfCon loss has the anchored function of $\mF$ and $\mG$, we obtain an upper bound of two symmetric mutual information. Here, $\vz_1=\vx$ because SelfCon loss is defined on the single-viewed batch and we assume that other images in a batch (i.e., $\vz_2, \cdots, \vz_K$) are sampled from the different class label with $\vx$.

\begin{align}
    \mathcal{I}(\mF(\vx);&~\mG(\vx)|c) + \mathcal{I}(\mG(\vx); \mF(\vx)|c) \\
    \geq& ~\mathbb{E} \log \frac{ e^{\psi_{\mF\mG}(\vx,\vx)}}{\frac{1}{2K-1} \big( e^{\psi_{\mF\mG}(\vx,\vx)} + \sum_{k=2}^{K} e^{\psi_{\mF\mG}(\vx,\vz_k)} + \sum_{k=2}^{K} e^{\psi_{\mF}(\vx,\vz_k)} \big)} \nonumber \\
    &+ \mathbb{E} \log \frac{ e^{\psi_{\mG\mF}(\vx,\vx)}}{\frac{1}{2K-1} \big( e^{\psi_{\mG\mF}(\vx,\vx)} + \sum_{k=2}^{K} e^{\psi_{\mG\mF}(\vx,\vz_k)} + \sum_{k=2}^{K} e^{\psi_{\mG}(\vx,\vz_k)} \big)} \\
    =& ~ 2\log(2K-1) - 2\gL_{self}(\vx;\{\mF,\mG\},K)
\end{align}

Due to the symmetry of mutual information,

\begin{align}\label{eq:placeholder5}
    \mathcal{I}(\mF(\vx);\mG(\vx)|c) \geq \log (2K-1) - \gL_{self}(\vx;\{\mF,\mG\},K)
\end{align}

\end{proof}

In addition, we can similarly bound the SelfCo loss with a multi-viewed batch (SelfCon-M). As the derivation of SupCon loss bound, only consider the anchor $\vx$ and its positive pair $\vz_1$. When the anchored feature is $\mF(\vx)$, the contrastive features are: $\mG(\vx)$, $\mG(\vz)$, and $\mF(\vz)$. By symmetry, when the anchored feature is $\mG(\vx)$, the contrastive features are: $\mF(\vx)$, $\mF(\vz)$, and $\mG(\vz)$. As the derivation of the SupCon loss bound, we assume the augmented negative samples, i.e., $\{\vz_2,\cdots,\vz_K,\vz_{K+1},\cdots,\vz_{2K-1}\}$.

\begin{align}
    \mathcal{I}(\mF(\vx);&~\mG(\vx)|c) + \mathcal{I}(\mF(\vx); \mG(\vz)|c) + \mathcal{I}(\mF(\vx); \mF(\vz)|c) \nonumber \\
    +\mathcal{I}(\mG(\vx);&~\mF(\vx)|c) + \mathcal{I}(\mG(\vx); \mF(\vz)|c) + \mathcal{I}(\mG(\vx); \mG(\vz)|c) \\
    \geq& ~ \frac{1}{3}~\mathbb{E} \log \frac{ e^{\psi_{\mF\mG}(\vx,\vx)} \cdot e^{\psi_{\mF\mG}(\vx,\vz_1)} \cdot e^{\psi_{\mF}(\vx,\vz_1)}}{\frac{1}{4K-1} \big( e^{\psi_{\mF\mG}(\vx,\vx)} + \sum_{k=1}^{2K-1} e^{\psi_{\mF\mG}(\vx,\vz_k)} + \sum_{k=1}^{2K-1} e^{\psi_{\mF}(\vx,\vz_k)} \big)} \nonumber \\
    &+ \frac{1}{3}~\mathbb{E} \log \frac{ e^{\psi_{\mG\mF}(\vx,\vx)} \cdot e^{\psi_{\mG\mF}(\vx,\vz_1)} \cdot e^{\psi_{\mG}(\vx,\vz_1)}}{\frac{1}{4K-1} \big( e^{\psi_{\mG\mF}(\vx,\vx)} + \sum_{k=1}^{2K-1} e^{\psi_{\mG\mF}(\vx,\vz_k)} + \sum_{k=1}^{2K-1} e^{\psi_{\mG}(\vx,\vz_k)} \big)} \\
    =& ~ \frac{2}{3}~\log(4K-1) - 2\gL_{self\text{-}m}(\vx,\vz;\{\mF,\mG\},K)
\end{align}

There could be a doubt about the loose bound between SelfCon loss and MI. However, when we prove the MI bound, we assumed a probabilistic model (refer to Eq. \ref{eq:placeholder10}). When the anchor feature is similar to the negative pairs (i.e., different class representations), this model becomes a variational distribution with random mapping, and SelfCon loss cannot be optimized at all. Therefore, optimizing SelfCon loss means that the representations of different classes get farther. Then, a better estimation of variational distribution leads to a small gap between SelfCon loss and MI. After all, SelfCon loss has improved performance because it tightens the bound of the label-conditional MI while distinguishing different class representations.

\subsection{Proof of Proposition \ref{prop:intermediate_feature}}

$\mF(\vx)$ and $\mG(\vx)$ are the output features from the backbone network and the sub-network, respectively. Recall that $\mT$ denotes the sharing layers between $\mF$ and $\mG$. $\mT(\vx)$ is the intermediate feature of the backbone, which is also an input to the auxiliary network path. 

Before proving Proposition \ref{prop:intermediate_feature}, we would like to note that the usefulness of mutual information should be carefully discussed on the stochastic mapping of a neural network. If a mapping $\mT(\vx) \mapsto \mF(\vx)$ is a deterministic mapping, then the MI between $\mT(\vx)$ and $\mF(\vx)$ is degenerate because $\gI(\mT(\vx);\mF(\vx))$ is either infinite for continuous $\mT(\vx)$ (conditional differential entropy is $-\infty$) or a constant for discrete $\mT(\vx)$ which is independent on the network’s parameters (equal to $\gH(\mT(\vx))$). However, for studying the usefulness of mutual information in a deep neural network, the map $\mT(\vx) \mapsto \mF(\vx)$ is considered as a stochastic parameterized channel. In many recent works about information theory with DNN, they view the training via SGD as a stochastic process, and the stochasticity in the training procedure lets us define the MI with stochastically trained representations \citep{opening_black_box,goldfeld2019estimating,saxe2019information,goldfeld2019estimating}. Our theoretical claim focuses on the SelfCon loss as a \textit{training} loss optimized by the SGD algorithm. Therefore, analyzing the MI between the hidden representations while training with the SelfCon loss is based on the information theory to understand DNN \citep{information_bottleneck}.

Also, information theory in deep learning, especially in contrastive learning, is based on the InfoMax principle \citep{linsker1989application} which is about learning a neural network that maps a set of input to a set of output to maximize the average mutual information between the input and output of a neural network, subject to stochastic processes. This InfoMax principle is nowadays widely used for analyzing and optimizing DNNs. Most works for contrastive learning are based on maximizing mutual information grounds on the InfoMax principle, and they are grounded on the stochastic mapping of an encoder. Moreover, \citet{interpolation_nce_nwj} rigorously discussed the mutual information with respect to a stochastic encoder. This is common practice in a representation learning context where $\vx$ is data, and $\vz$ is a learned stochastic representation.

\begin{equation}\label{eq:placeholder11}
\gI(\mF(\vx);\mG(\vx)|c) \leq \gI(\mF(\vx);\mT(\vx)|c)    
\end{equation}

\begin{proof}
As $\mF(\vx)$ and $\mG(\vx)$ are conditionally independent given the intermediate representation $\mT(\vx)$, they formulate a Markov chain as follows: $\mG \leftrightarrow \mT \leftrightarrow \mF$ \citep{markov_chain}. Under this relation, the following is satisfied:

\begin{align}
\label{eq:conditional entropy}
    \gI(\mF(\vx);\mG(\vx)|c)
    & = \gH(\mF(\vx)|c) - \gH(\mF(\vx)|\mG(\vx),c) \\ 
    & \leq \gH(\mF(\vx)|c) - \gH(\mF(\vx)|\mT(\vx),\mG(\vx),c) \label{eq:placeholder3} \\ 
    & = \gH(\mF(\vx)|c) - \int_{\mathbf{t}, \mathbf{f}, \mathbf{g}} p(\mathbf{t}, \mathbf{f}, \mathbf{g}|c) \log p(\mathbf{f}|\mathbf{t}, \mathbf{g}, c) d\mathbf{t}d\mathbf{f}d\mathbf{g} \\
    & = \gH(\mF(\vx)|c) - \int_{\mathbf{t}, \mathbf{f}} p(\mathbf{t}, \mathbf{f}|c) \log p(\mathbf{f}|\mathbf{t}, c) d\mathbf{t}d\mathbf{f} \label{eq:placeholder4} \\
    & = \gH(\mF(\vx)|c) - \gH(\mF(\vx)|\mT(\vx), c) \\
    & = \gI(\mF(\vx);\mT(\vx)|c)
\end{align}

\noindent Eq. \ref{eq:placeholder3} is from the property of conditional entropy, and Eq. \ref{eq:placeholder4} is due to the conditional independence and marginalization of $\mathbf{g}$. 
\end{proof}

From Eq. \ref{eq:placeholder11} and Eq. \ref{eq:placeholder5}, we further obtain Eq. \ref{eq:decomposed_mi} as follows:

\begin{align}
    \log (2K-1) &-\mathcal{L}_{self\text{-}s}(\vx;\{\mF,\mG\},K) \\
    & \leq \gI(\mF(\vx);\mG(\vx)|c) \\
    & \leq \gI(\mF(\vx);\mT(\vx)|c) \\ 
    & = \gI(\mF(\vx);\mT(\vx),c) - \gI(\mF(\vx);c) \\
    & = \underbrace{\gI(\mF(\vx);\mT(\vx))}_{(\square)} + \underbrace{\gI(\mF(\vx);c|\mT(\vx)) - \gI(\mF(\vx);c)}_{(\blacksquare)} \label{eq:placeholder8}
\end{align}

Strictly speaking, SelfCon loss does not guarantee the lower bound of either ($\square$) or ($\blacksquare$) in Eq. \ref{eq:placeholder8}. However, SelfCon loss guarantees the label-conditional MI between the intermediate and the last feature, which is ($\square + \blacksquare$). 



\newpage

\section{Implementation Details}
\label{app:implementation details}

\subsection{Network Architectures}

We modified the architecture of networks according to the benchmarks. For the smaller scale of benchmarks (e.g., CIFAR-10, CIFAR-100, and Tiny-ImageNet) and the residual networks (e.g., ResNet-18, ResNet-50, and WRN-16-8), we changed the kernel size and stride of a convolution head to 3 and 1, respectively. We also excluded Max-Pooling on the top of the ResNet architecture for the CIFAR datasets. Moreover, for VGG-16 with BN, the dimension of the fully-connected layer was changed from 4096 to 512 for CIFAR and Tiny-ImageNet. MLP projection head for contrastive learning consisted of two convolution layers with 128 dimensions and one ReLU activation. For the architectures of sub-networks, refer to Appendix \ref{app:ablation_subnet}.

\subsection{Representation Learning}

We refer to the technical improvements used in SupCon, i.e., a cosine learning rate scheduler \citep{cosine_schedule}, an MLP projection head \citep{simclr}, and the augmentation strategies \citep{autoaugment}: \{ResizedCrop, HorizontalFlip, ColorJitter, GrayScale\}. ColorJitter and GrayScale are only used in the pretraining stage. For small-scale benchmarks, we used 8 GPUs and set the batch size as 1024 for the pretraining and 512 for the linear evaluation. We trained the encoder and the linear classifier for 1000 epochs and 100 epochs, respectively. 
For large-scale benchmarks, in Table \ref{tab:imagenet_total}, we used batch size of 512 when pretraining on the ImageNet-100 benchmark. Besides, we trained the encoder and the linear classifier for 400 epochs and 40 epochs, respectively. For ImageNet benchmark, we used batch size of 2048, except for the ResNet-50 experiments, where we used batch size of 1024 due to the limited memory capacity. We trained the encoder and the linear classifier for 800 epochs and 40 epochs, respectively.

Every experiment used SGD with 0.9 momentum and weight decay of 1e-4 without Nesterov momentum. All contrastive loss functions used temperature $\tau$ of 0.1. For a fair comparison to \citet{supcon}, we set the same learning rate of the encoder network as 0.5 for the small-scale benchmarks. Refer to Appendix \ref{app:ablation_subnet} for the optimal learning rate of the large-scale dataset. We linearly scaled the learning rate according to the batch size \citep{lr_batch}. On the linear evaluation step, we used 5.0 as a learning rate of the linear classifier for the residual architecture, but it was robust to any value and converged in nearly 20 epochs. Meanwhile, for VGG architecture, only a small learning rate of 0.1 converged.

\subsection{Datasets for Downstream Tasks}
For the fine-grained recognition task datasets, we used CUB \citep{wah2011caltech}, Stanford Dogs \citep{khosla2011novel}, MIT67 \citep{quattoni2009recognizing}, Flowers \citep{nilsback2008automated}, Pets \citep{parkhi2012cats}, Stanford40 \citep{yao2011human}, Stanford Cars \citep{krause2013collecting}, and FGVC Aircraft \citep{maji2013fine}. Also, for the semantic segmentation and object detection task datasets, we used Pascal VOC \citep{everingham2010pascal} and MS COCO \citep{lin2014microsoft}.
Especially, in the object detection, the evaluation metrics for Pascal VOC 2007 and MS COCO datasets are mAP@0.5 and mAP@[.50:.05:.95], respectively, following the conventional evaluation protocol. In case of the Pascal VOC detection, we used the trainval set of Pascal VOC 2007 (5,011 images) for training and the test set (4,952 images) for evaluation.

\subsection{Hyperparameters}\label{app:ablation_subnet}

\paragraph{Sensitivity study for learning rate}


In Table \ref{tab:smaller_batch}, we experimented with the supervised contrastive algorithms with various batch sizes and confirmed that the classification accuracy decreases in the large batch size. We supposed that this trend is induced by the regularization effect from the batch size. However, there could be a concern for using a sub-optimal learning rate on the large batch size.

We further studied the sensitivity for the learning rate in a batch size of 1024 on CIFAR-100 and summarized the results in Table \ref{tab:sensitivity_study}. We concluded that the performance comparison in Table \ref{tab:smaller_batch} is consistent with hyperparameter tuning. The experimental results supported that a larger learning rate than 0.5 may be a better choice but the trend between all methods maintained in parallel with the learning rate of 0.5. Therefore, we stick to the initial learning rate of 0.5 that \citet{supcon} had used.

Moreover, we tuned the learning rate for the reliability of our large-scale experiments in ImageNet-100 and ImageNet datasets. Table \ref{tab:learning_rate_imagenet100} summarizes the sensitivity results in ImageNet-100 on ResNet-18 architecture. In this experiment, we fixed the batch size to 512 and pretraining epochs to 400. Note that although we used the \textit{same} sub-network in this experiment, Table \ref{tab:imagenet_total} reports the results with the \textit{small} sub-network after our sub-network experiments in the next section. We confirmed that the learning rate of 0.5 is the best except for SupCon, which showed the best at 1.0. Every other experiment in ImageNet-100 used this best setting.

Table \ref{tab:learning_rate_imagenet} summarizes the results in ImageNet on ResNet-18 architecture. We fixed the batch size to 3072 and pretraining epochs to 400 for the fast experiments. We compared SelfCon with SupCon, two major methods in our paper, and confirmed that the learning rate of 0.375 is the best. We used 0.375 for other experiments, and we used a linear scaling rule for other batch sizes (e.g., 0.25 for 2048 batch size or 0.125 for 1024 batch size). 

\begin{table}[!t]
\vspace{-2pt}
    \begin{subfigure}[t]{0.35\linewidth}
    \centering
    \small
    \addtolength{\tabcolsep}{-2pt}
    \resizebox{\linewidth}{!}{
        \begin{tabular}{lcccccc}
        \toprule
        & \multicolumn{6}{c}{Learning Rate} \\
        \cmidrule(l{2pt}r{2pt}){2-7}
        \multirow{-2.5}{*}{Method} & $0.125$ & $0.25$ & $0.5$ & $1.0$ & $2.0$ & $4.0$ \\
        \midrule\hline
        SupCon & $71.8$ & $72.3$ & $73.0$ & $73.4$ & $74.6$ & $\mathbf{74.7}$ \\
        SelfCon-M & $72.3$ & $73.6$ & $74.9$ & $\mathbf{75.5}$ & $\mathbf{75.5}$ & $75.1$ \\
        SupCon-S & $72.3$ & $73.1$ & $73.9$ & $74.6$ & $\mathbf{74.8}$ & $73.6$ \\
        SelfCon &  $74.3$ & $74.6$ & $75.4$ & $75.7$ & $\mathbf{76.0}$ & $74.7$ \\ \bottomrule
        \end{tabular}}
    \caption{CIFAR-100}
    \label{tab:sensitivity_study}
    \end{subfigure}%
    \hfill
    \begin{subfigure}[t]{0.35\linewidth}
    \centering
    \small
    \addtolength{\tabcolsep}{-2pt}
    \resizebox{\linewidth}{!}{
        \begin{tabular}{lcccccc}
        \toprule
        & \multicolumn{6}{c}{Learning Rate} \\
        \cmidrule(l{2pt}r{2pt}){2-7}
        \multirow{-2.5}{*}{Method} & $0.125$ & $0.25$ & $0.5$ & $1.0$ & $2.0$ & $4.0$ \\
        \midrule\hline
        SupCon & $83.4$ & $84.5$ & $85.2$ & $\mathbf{85.6}$ & $84.3$ & $75.4$ \\
        SelfCon-M & $84.6$ & $85.2$ & $\mathbf{85.8}$ & $85.5$ & $84.8$ & $73.4$ \\  
        SupCon-S & $83.8$ & $84.7$ & $\mathbf{84.9}$ & $84.8$ & $82.3$ & $1.0$ \\
        SelfCon &  $84.8$ & $85.4$ & $\mathbf{85.7}$ & $85.4$ & $84.0$ & $56.3$ \\ \bottomrule
        \end{tabular}}
    \caption{ImageNet-100}
    \label{tab:learning_rate_imagenet100}
    \end{subfigure}%
    \hfill
    \begin{subfigure}[t]{0.27\linewidth}
    \centering
    \small
    \addtolength{\tabcolsep}{-1pt}
    \resizebox{0.95\linewidth}{!}{
    \begin{tabular}{lcccc}
    \toprule
    & \multicolumn{4}{c}{Learning Rate} \\
    \cmidrule(l{2pt}r{2pt}){2-5}
    \multirow{-2.5}{*}{Method} & $0.1875$ & $0.375$ & $0.75$ & $1.5$ \\
    \midrule\hline
    SupCon & $71.0$ & $\mathbf{71.2}$ & $\mathbf{71.2}$ & $70.7$ \\
    SelfCon &  $\mathbf{71.3}$ & $\mathbf{71.3}$ & $71.1$ & $69.1$ \\ \bottomrule
    \end{tabular}}
    \caption{ImageNet}
    \label{tab:learning_rate_imagenet}
    \end{subfigure}%
    \vspace{-8pt}
    \caption{Classification accuracy on ResNet-18 with various learning rates. We used a fully-connected layer (to be defined as \textit{fc}) as the sub-network for SelfCon methods on CIFAR-100. For the other benchmarks, non-sharing blocks of the backbone and sub-network are the same (to be defined as \textit{same}).}
    \label{tab:learning_rate}
\end{table}

\paragraph{Sub-network}
The structure, position, and number of sub-networks are important to the performance of SelfCon learning.
First, in order to find a suitable structure of the sub-network, the following three structures were attached after the $2^\text{nd}$ block of an encoder: (1) a simple \textit{fc}, fully-connected, layer, (2) \textit{small} structure which reduced by half the number of layers in the non-sharing blocks, (3) \textit{same} structure which is same as the backbone's non-sharing block structure. After we found the optimal structure, we fixed the structure of the sub-network and found which position was the best. For ResNet architectures, there are three positions to attach; after the $1^\text{st}$, $2^\text{nd}$, and $3^\text{rd}$ block. For VGG-16 with BN, there are four positions, and for WRN-16-8, there are two positions possible. Note that blocks are divided based on the Max-Pooling layer in VGG-16 with BN. 

Table \ref{tab:ablation_resnet18} presents the ablation study results for ResNet-18 and ResNet-50, and Table \ref{tab:ablation_wrn} presents the results for WRN-16-8 and VGG-16 with BN. We observed a trend: in shallow networks (e.g., WRN-16-8) the \textit{same} structure was better, while \textit{fc} was better in deeper networks (e.g., VGG-16 with BN and ResNet). Besides, the performance was consistently good when the exit path is attached after the midpoint of the encoder (e.g., $2^\text{nd}$ block in ResNet or $3^\text{rd}$ block in VGG architecture). 

\begin{table}[!t]
\begin{subfigure}[b]{0.45\linewidth}
\raggedleft
\scriptsize
\setlength{\tabcolsep}{8pt}
\renewcommand{\arraystretch}{0.9}
\begin{tabular}{lcccc}
\toprule
    & \multicolumn{3}{c}{Position} & \\ \cmidrule(l{2pt}r{2pt}){2-4}
 \multirow{-2.5}{*}{Structure} & $1^{\text{st}}$ Block    &  $2^{\text{nd}}$ Block    &   $3^{\text{rd}}$ Block  & \multirow{-2.5}{*}{Accuracy} \\ \midrule \hline
     FC   &   &   \cmark    &   &  $\mathbf{75.4_{\pm{0.1}}}$ \\
     Small   &   &  \cmark      &   &  $74.7_{\pm{0.2}}$ \\
     Same   &   &  \cmark      &   &  $74.5_{\pm{0.0}}$ \\ \hline
     FC   & \cmark     &    &   &  $73.2_{\pm{0.2}}$ \\
     FC   &   &  \cmark  &   &  $\mathbf{75.4_{\pm{0.1}}}$ \\
     FC   &   &   & \cmark    &  $\mathbf{75.5_{\pm{0.1}}}$ \\
     FC   & \cmark  &  \cmark  & \cmark  &  $74.5_{\pm{0.1}}$ \\ \bottomrule
\end{tabular}
\end{subfigure}
\hfill%
\begin{subfigure}[b]{0.45\linewidth}
\raggedright
\scriptsize
\setlength{\tabcolsep}{8pt}
\renewcommand{\arraystretch}{0.9}
\begin{tabular}{lcccc}
\toprule
    & \multicolumn{3}{c}{Position} & \\ \cmidrule(l{2pt}r{2pt}){2-4}
 \multirow{-2.5}{*}{Structure} & $1^{\text{st}}$ Block    &  $2^{\text{nd}}$ Block    &   $3^{\text{rd}}$ Block  & \multirow{-2.5}{*}{Accuracy} \\ \midrule \hline
     FC   &   &   \cmark    &   &  $\mathbf{78.5_{\pm{0.3}}}$ \\
     Small   &   &  \cmark      &   &  $\mathbf{78.1_{\pm{0.2}}}$ \\
     Same   &   &  \cmark      &   &  $77.4_{\pm{0.2}}$ \\ \hline
     FC   & \cmark     &    &   &  $77.0_{\pm{0.2}}$ \\
     FC   &   &  \cmark  &   &  $\mathbf{78.5_{\pm{0.3}}}$ \\
     FC   &   &   & \cmark    &  $77.4_{\pm{0.1}}$ \\
     FC   & \cmark  &  \cmark  & \cmark  &  $\mathbf{78.7_{\pm{0.5}}}$ \\ \bottomrule
\end{tabular}
\end{subfigure}
\vspace{-5pt}
\caption{The results of SelfCon loss according to the structure and position of sub-network. The classification accuracy is for ResNet-18 (Left) and ResNet-50 (Right) on the CIFAR-100 benchmark.}
\label{tab:ablation_resnet18}
\end{table}

\begin{table}[!t]
\begin{subfigure}[b]{0.4\linewidth}
\raggedleft
\scriptsize
\setlength{\tabcolsep}{10pt}
\renewcommand{\arraystretch}{1.1}
\begin{tabular}{lccc}
\toprule
    & \multicolumn{2}{c}{Position} & \\ \cmidrule(l{2pt}r{2pt}){2-3}
 \multirow{-2.5}{*}{Structure} & $1^{\text{st}}$ Block    &  $2^{\text{nd}}$ Block   & \multirow{-2.5}{*}{Accuracy} \\ \midrule \hline
     FC   &  \cmark &    &  $74.4_{\pm{1.2}}$ \\
     Small   & \cmark  &    &  $76.2_{\pm{0.0}}$ \\
     Same   &  \cmark &    & $\mathbf{76.6_{\pm{0.1}}}$ \\ \hline
     Same   & \cmark   &   &  $\mathbf{76.6_{\pm{0.1}}}$ \\
     Same   &   &  \cmark  & $\mathbf{76.5_{\pm{0.2}}}$ \\
     Same   & \cmark  & \cmark    &  $76.5_{\pm{0.0}}$ \\
     \bottomrule
\end{tabular}
\end{subfigure}
\hfill
\begin{subfigure}[b]{0.53\linewidth}
\raggedright
\scriptsize
\setlength{\tabcolsep}{8pt}
\renewcommand{\arraystretch}{0.88}
\begin{tabular}{lccccc}
\toprule
    & \multicolumn{4}{c}{Position} & \\ \cmidrule(l{2pt}r{2pt}){2-5}
 \multirow{-2.5}{*}{Structure} & $1^{\text{st}}$ Block    &  $2^{\text{nd}}$ Block    &   $3^{\text{rd}}$ Block & $4^{\text{th}}$ Block  & \multirow{-2.5}{*}{Accuracy} \\ \midrule \hline
     FC   &   &   \cmark &    &   &  $\mathbf{71.4_{\pm{0.0}}}$ \\
     Small   &   &  \cmark  &    &   &  $\mathbf{71.5_{\pm{0.4}}}$ \\
     Same   &   &  \cmark   &   &   &  $\mathbf{71.5_{\pm{0.3}}}$ \\ \hline
     FC   & \cmark &    &  &   &  $70.9_{\pm{0.1}}$ \\
     FC   &   &  \cmark  &   & & $71.4_{\pm{0.0}}$ \\
     FC   &   &   & \cmark &   &  $72.0_{\pm{0.0}}$ \\
     FC   &   &   &  & \cmark  &  $71.5_{\pm{0.1}}$ \\
     FC   & \cmark  &  \cmark  & \cmark  & \cmark & $\mathbf{72.5_{\pm{0.1}}}$ \\ \bottomrule
\end{tabular}
\end{subfigure}
\vspace{-5pt}
\caption{The results of SelfCon loss according to the structure and position of sub-network. The classification accuracy is for WRN-16-8 (Left) and VGG-16 with BN (Right) on the CIFAR-100 benchmark.}
\vspace{-10pt}
\label{tab:ablation_wrn}
\end{table}

Obviously, there are many combinations of placing sub-networks, and Table \ref{tab:ablation_resnet18} and Table \ref{tab:ablation_wrn} presented an interesting result that some performance was the best when sub-networks are attached to all blocks.
It seems that increasing the number of positive and negative pairs by various views from multiple sub-networks improves the performance.
It is consistent with the argument of CMC \citep{cmc} that the more views, the better the representation, but our SelfCon learning is much more efficient in terms of the computational cost and GPU memory usage.
However, for the efficiency of the experiments and a better understanding of the framework, we stuck to a single sub-network in all experimental settings. 


We further experimented on the large-scale benchmarks, ImageNet-100 and ImageNet. In the ImageNet-100 experiment, we fixed the learning rate for SelfCon and SelfCon-M to 0.5, which was found to be optimal in the previous section.
SelfCon showed the accuracy of 85.5$\%$, 86.1$\%$, and 85.7$\%$ in the order of \textit{fc}, \textit{small}, and \textit{same} structures, respectively, and in the case of SelfCon-M, the results were 84.6$\%$, 85.8$\%$, and 85.8$\%$, respectively.
We found out that \textit{small} sub-network showed the best performance in SelfCon as well as SelfCon-M. Therefore, we fixed the \textit{small} sub-network with a learning rate of 0.5 for the SelfCon methods.
For ImageNet dataset, we compared the results of \textit{small} structure with those of \textit{same} structure. The classification performance degraded by 0.3\% when we used \textit{small} structure. Therefore, we concluded that a deeper sub-network structure is preferred in the large-scale benchmark.

\newpage
\section{Extensions of SelfCon Learning}
\label{app:extensions}

\subsection{SelfCon in Unsupervised Learning}
\label{app:self_sup}

Although we have experimented only in supervision, our motivation of contrastive learning with a multi-exit framework can also be extended to unsupervised learning.
We propose a SelfCon loss function for the unsupervised scenario and present the linear evaluation performance of ResNet-18 architecture on the CIFAR-100 dataset.


\paragraph{Loss function}

Under the unsupervised setting, \citet{simclr} proposed a simple framework for contrastive learning of visual representations (SimCLR) with NT-Xent loss.   
SimCLR suggests a contrastive task that contrasts the augmented pair among other images.
Therefore, the objective of SimCLR is exactly same as Eq. \ref{eq:supcon}, while each sample has only one positive pair of its own augmented image, i.e., $P_i \equiv \{ (i + B) \bmod 2B \}$. We denote this loss as $\gL_{sim}$.

We formulate SelfCon loss in an unsupervised setting as SimCLR, using a positive set without label information. We formulate \textbf{SelfCon} loss with the \textbf{s}ingle-viewed and \textbf{u}nlabeled batch (\textbf{SelfCon-SU}) as follows:
\vspace{5pt}
\begin{gather}
  \mathcal{L}_{self\text{-}su} = -\sum_{\substack{ \textcolor{anchor}{i} \in I,\\\bm{\omega} \in \bm{\Omega}}} \frac{1}{|P_{i1}||\bm{\Omega}|} \sum_{\substack{ \textcolor{pos}{p_1} \in P_{i1}, \\\bm{\omega}_1 \in \bm{\Omega}}} \log \frac{ \exp(\bm{\omega}(\textcolor{anchor}{\vx_i})^\top \bm{\omega}_1(\textcolor{pos}{\vx_{p_1}}))}{\!\!\sum\limits_{\bm{\omega}_2 \in \bm{\Omega}}\!\!\bigg( \sum\limits_{\textcolor{pos}{p_2} \in P_{i2}} \!\! \exp(\bm{\omega}(\textcolor{anchor}{\vx_i})^\top \bm{\omega}_2(\textcolor{pos}{\vx_{p_2}})) +\!\! \sum\limits_{\textcolor{neg}{n} \in N_i}\!\! \exp(\bm{\omega}(\textcolor{anchor}{\vx_i})^\top \bm{\omega}_2(\textcolor{neg}{\vx_n}))\bigg)} \nonumber \\
    \label{eq:selfcon_su}
    \boxed{I \equiv \{1,\dots,B\}} \,\,\,
    \boxed{P_{ij} \equiv \{i\} } \,\,\,
    \boxed{N_i \equiv I \setminus \{i\}}
\vspace{10pt}
\end{gather}
We exclude the sample from the positive set only when $\bm{\omega} = \bm{\omega}_j$, i.e., $P_{ij} \leftarrow P_{ij} \setminus \{i\}$. For the positive set, since this loss is based on the single-viewed batch, we have an empty positive set when $\bm{\omega} = \bm{\omega}_j$. Here, we used $\tau=0.5$ for the unsupervised SelfCon loss and $\gL_{sim}$. We omitted the dividing constant for the summation of anchor samples (i.e., $(|I||\bm{\Omega}|)^{-1}$).

We also formulate \textbf{SelfCon} loss with the \textbf{m}ulti-viewed and \textbf{u}nlabeled batch (\textbf{SelfCon-MU}) as follows:
\vspace{5pt}
\begin{gather}
  \mathcal{L}_{self\text{-}mu} = -\sum_{\substack{ \textcolor{anchor}{i} \in I,\\\bm{\omega} \in \bm{\Omega}}} \frac{1}{|P_{i1}||\bm{\Omega}|} \sum_{\substack{ \textcolor{pos}{p_1} \in P_{i1}, \\\bm{\omega}_1 \in \bm{\Omega}}} \log \frac{ \exp(\bm{\omega}(\textcolor{anchor}{\vx_i})^\top \bm{\omega}_1(\textcolor{pos}{\vx_{p_1}}))}{\!\!\sum\limits_{\bm{\omega}_2 \in \bm{\Omega}}\!\!\bigg( \sum\limits_{\textcolor{pos}{p_2} \in P_{i2}} \!\! \exp(\bm{\omega}(\textcolor{anchor}{\vx_i})^\top \bm{\omega}_2(\textcolor{pos}{\vx_{p_2}})) +\!\! \sum\limits_{\textcolor{neg}{n} \in N_i}\!\! \exp(\bm{\omega}(\textcolor{anchor}{\vx_i})^\top \bm{\omega}_2(\textcolor{neg}{\vx_n}))\bigg)} \nonumber \\
  \label{eq:selfcon_mu}
    \boxed{I \equiv \{1,\dots,2B\}} \,\,\,
    \boxed{P_{ij} \equiv \{i, (i + B) \bmod 2B \}} \,\,\,
    \boxed{N_i \equiv I\setminus \{i, (i+B) \bmod 2B \}}
\vspace{10pt}
\end{gather}
Similarly, we exclude an anchor sample from the positive set, i.e., $P_{ij}\leftarrow P_{ij} \setminus \{i\}$ when $\bm{\omega} = \bm{\omega}_j$.

\paragraph{Experimental results}
All implementation details for unsupervised representation learning are identical with those of supervised representation learning in Appendix \ref{app:implementation details}, except for temperature $\tau$ of 0.5 and linear evaluation learning rate of 1.0. We used a \textit{small} sub-network attached after the $2^{\text{nd}}$ block.
Table \ref{tab:unsupervised} shows the linear evaluation performance of unsupervised learning on ResNet-18 in CIFAR-100 dataset. However, we empirically found that the encoder failed to converge with both SelfCon-SU and SelfCon-MU loss (see the accuracy of the first row).

\begin{table}[!ht]
\small
\centering
\begin{tabular}{lcccc}
\toprule
Method & CE & SimCLR & SelfCon-MU & SelfCon-SU \\
\midrule
Multi-view & - & \cmark & \cmark & \xmark \\
\midrule \hline
Accuracy & $72.9_{\pm 0.1}$ & $63.3_{\pm 0.3}$ & $5.0 _{\pm 0.1}$ & $6.4_{\pm 0.2}$ \\
\textbf{Accuracy*} & - & - & $\mathbf{64.6_{\pm 0.1}}$ & $12.8_{\pm 0.1}$ \\
\bottomrule
\end{tabular}
\caption{The results under the unsupervised scenario. We compared our SelfCon-SU and SelfCon-MU loss with SimCLR in the unsupervised setting. For the comparison with supervised learning, we also added the classification accuracy of CE loss. We used ResNet-18 encoder and CIFAR-100 dataset. Accuracy* denotes the accuracy of SelfCon learning with the anchors only from the sub-network (see details in Appendix \ref{app:anchor_only_subnet}).}
\label{tab:unsupervised}
\end{table}

\subsection{SelfCon with Anchors ONLY from the Sub-network}\label{app:anchor_only_subnet}

We suspect that Eq. \ref{eq:selfcon_su} and \ref{eq:selfcon_mu} allow the backbone network to follow the sub-network, which makes the last feature learn more redundant information about the input variable without any label information. Thus, the unsupervised loss function under the SelfCon framework needs to be modified. 

When the anchor feature is from the backbone network, we remove the loss term, which contrasts the features of the sub-network. Strictly speaking, it does not perfectly prevent the backbone from following the sub-network because there is no stop-gradient operation on the outputs of the backbone network when the outputs of the sub-network are the anchors. However, we hypothesize that it helps prevent the encoder from collapsing to the trivial solution by the contradiction of the IB principle. We confirmed the performance of revised loss functions in both unsupervised and supervised scenarios.

\paragraph{Loss function}

\begin{gather}
  \mathcal{L}_{self\text{-}su\text{*}} = - \sum_{\textcolor{anchor}{i} \in I} \frac{1}{|P_{i1}|} \sum_{\textcolor{pos}{p_1} \in P_{i1}} \!\log \frac {\exp(\mG(\textcolor{anchor}{\vx_i})^\top \mF(\textcolor{pos}{\vx_{p_1}}))}{\!\! \sum\limits_{\textcolor{pos}{p_2} \in P_{i2}} \!\! \exp(\mG(\textcolor{anchor}{\vx_i})^\top \mF(\textcolor{pos}{\vx_{p_2}})) +\!\! \sum\limits_{\substack{\textcolor{neg}{n} \in N_i,\\\bm{\omega}_2 \in \bm{\Omega}}}\!\! \exp(\mG(\textcolor{anchor}{\vx_i})^\top \bm{\omega}_2(\textcolor{neg}{\vx_n}))} \nonumber \\
    \boxed{I \equiv \{1,\dots,B\}} \,\,\,
    \boxed{P_{ij} \equiv \{i\}} \,\,\,
    \boxed{N_i \equiv I \setminus \{i\}}
  \label{eq:selfcon-su without fg}
\end{gather}

\noindent All notations are same as Eq. \ref{eq:selfcon_su}. For the supervised setting, we change the above equation as the equally-weighted linear combination of SupCon-S loss and Eq. \ref{eq:selfcon-su without fg} with $P_{ij} \equiv \{ p_j \in I | y_p = y_i \}$. Note that we also exclude contrasting the anchor itself in SupCon-S loss term.

\begin{align}
    &\mathcal{L}_{self\text{-}mu\text{*}} = \frac{1}{1+\alpha} \mathcal{L}_{sim} \,+  \nonumber \\
    &\frac{\alpha}{1+\alpha} \Bigg[ - \sum_{\textcolor{anchor}{i} \in I} \frac{1}{|P_{i1}||\bm{\Omega}|} \sum_{\substack{\textcolor{pos}{p_1} \in P_{i1},\\\bm{\omega}_1\in \bm{\Omega}}} \!\log \frac {\exp(\mG(\textcolor{anchor}{\vx_i})^\top \bm{\omega}_1(\textcolor{pos}{\vx_{p_1}}))}{\!\!\sum\limits_{\bm{\omega}_2 \in \bm{\Omega}}\!\!\bigg( \sum\limits_{\textcolor{pos}{p_2} \in P_{i2}} \!\! \exp(\mG(\textcolor{anchor}{\vx_i})^\top \bm{\omega}_2(\textcolor{pos}{\vx_{p_2}})) +\!\! \sum\limits_{\textcolor{neg}{n} \in N_i}\!\! \exp(\mG(\textcolor{anchor}{\vx_i})^\top \bm{\omega}_2(\textcolor{neg}{\vx_n}))\bigg)} \Bigg] \nonumber \\
    \label{eq:selfcon-mu without fg}
    &\hspace{20pt}\boxed{I \equiv \{1,\dots,2B\}} \,\,\,
    \boxed{P_{ij} \equiv \{(i + B) \bmod 2B \}} \,\,\,
    \boxed{N_i \equiv I\setminus \{ i, (i + B) \bmod 2B \}}
\end{align}
\vspace{5pt}

\noindent All notations are same as Eq. \ref{eq:selfcon_mu}, except for the coefficient $\alpha$ where we used 1.0. For the supervised setting, simply change $P_{ij}$ to $\{ p_j \in I \setminus \{i\} | y_p = y_i \}$ and $\mathcal{L}_{sim}$ to $\mathcal{L}_{sup}$. Note that $P_{ij}\leftarrow P_{ij}\cup\{i\}$ when $\bm{\omega}_j = \mF$. We get rid of the situation that the anchor $\mF(\bm{x})$ contrasts the positive pair in sub-network $\mG(\bm{x}_{p_j})$. Still, the proposed loss function includes SimCLR loss ($\gL_{sim}$) or SupCon loss ($\gL_{sup}$) in the unsupervised or supervised setting, respectively.

\paragraph{Experimental results}
In Table \ref{tab:unsupervised}, we also reported the accuracy of SelfCon-SU* and SelfCon-MU* loss according to Eq. \ref{eq:selfcon-su without fg} and \ref{eq:selfcon-mu without fg}. Surprisingly, in this case, SelfCon-MU* outperformed SimCLR loss \citep{simclr}, improving 1.3\%. 
It is a consistent result with the recent works, which boost the performance by directly formulating the contrastive task within the intermediate layers \citep{local_dim, loco, inter_moco}.
Unfortunately, SelfCon-SU* had not converged again, although it improved the result in a small amount compared to Eq. \ref{eq:selfcon_su}. 
While SelfCon-MU* has SimCLR loss term that makes the backbone encoder still learn meaningful features, SelfCon-SU* loss does not have the anchor features from the backbone, which makes the backbone hard to be trained.
Table \ref{tab:remove_fg} summarizes SelfCon* and SelfCon-M* loss, removing the anchors from the backbone in the supervised setting, i.e., supervised version of Eq. \ref{eq:selfcon-su without fg} and \ref{eq:selfcon-mu without fg}. As we expected, these variants of SelfCon and SelfCon-M further improved the classification performance.

\begin{table}[!ht]
\small
\centering
\begin{tabular}{lccc}
\toprule
Method & Architecture & Accuracy & \textbf{Accuracy*} \\
\midrule \hline
SelfCon-M & \multirow{2}{*}{ResNet-18} & $\mathbf{74.9_{\pm 0.1}}$ & $\mathbf{74.9_{\pm 0.2}}$ \\
SelfCon & & $75.4_{\pm 0.1}$ & $\mathbf{75.6_{\pm 0.1}}$ \\
\hline
SelfCon-M & \multirow{2}{*}{ResNet-50} & $76.9_{\pm 0.1}$ & $\mathbf{77.4_{\pm 0.2}}$ \\
SelfCon & & $\mathbf{78.5_{\pm 0.3}}$ & $\mathbf{78.8_{\pm 0.1}}$ \\
\bottomrule
\end{tabular}
\caption{CIFAR-100 results with SelfCon extensions. Accuracy* denotes the accuracy of SelfCon learning with the anchors only from the sub-network. We used $\alpha=1$ and a fully-connected layer as the sub-network structure.}
\vspace{-10pt}
\label{tab:remove_fg}
\end{table}

\clearpage
\section{1-Stage Training}
\label{app:1-stage_training}

\paragraph{Implementation details}
We experimented our SelfCon with a 1-stage training framework, i.e., not decoupling the encoder pretraining and fine-tuning. 
In the 1-stage training protocol, we trained the encoder network jointly with a linear classifier on the single-viewed batch. Most of the experimental settings were the same as those of representation learning, but we trained the encoder for 500 epochs with a cosine learning rate scheduler on all benchmarks except ImageNet. We used the batch size of 1024 and the learning rate of 0.8 for small-scale datasets, and 512 batch size and 0.4 learning rate for ImageNet-100. For the cross-entropy result of the ImageNet dataset, we trained for 90 epochs with the multi-step learning rate scheduler after 30 and 60 epochs with the decay ratio of 0.1.


\paragraph{Performance comparison}
For a fair comparison, we select baselines as the standard supervised methods with sub-network and 1-stage version of SelfCon.
In the multi-exit framework, we used a linear combination of loss functions for the backbone and sub-network. We used only cross-entropy loss for the backbone network and weighted linear combinations of the loss functions (e.g., KL divergence and SelfCon) for the sub-network. For example, Self-Distillation \citep{be_your_own} used the interpolation coefficient $\alpha$ of 0.5. For the 1-stage version of SelfCon loss, we follow the coefficient form in \citep{crd}: $\mathcal{L} = \mathcal{L}_{CE} + \beta \mathcal{L}_{self}$. We set the coefficient $\beta=0.8$ for all experiments. Note that we used the outputs from the projection head instead of the logits. We did not use the interpolated form, unlike SD, because we distill the features from the projection head. We used temperature $\tau=3.0$ for SD and $\tau=0.1$ for SelfCon loss.

From Table \ref{tab:1-stage}, we observed that simply adding cross-entropy loss to the sub-network (CE w/ Sub) improved the backbone network's classification performance. However, the results of SD suggest a saturation of the backbone's accuracy even when the classifier of the sub-network converges well.
SelfCon loss in 2-stage training still demonstrated the best classification accuracy against to the 1-stage training methods.
We note that various performance boosting techniques can also be applied to the fine-tuning phase of the 2-stage training.

\begin{table}[!h]
\vspace{5pt}
\centering
\scriptsize
\addtolength{\tabcolsep}{-4pt}
\resizebox{0.5\linewidth}{!}{
\renewcommand{\arraystretch}{0.8}
    \begin{tabular}{lcccc}
    \toprule
     & \multicolumn{2}{c}{ResNet-18} & \multicolumn{2}{c}{ResNet-50} \\ 
     \cmidrule(l{2pt}r{2pt}){2-3}\cmidrule(l{2pt}r{2pt}){4-5}
    \multirow{-2.5}{*}{Method} & CF-100 & IN-100 & CF-100 & IN-100 \\ \midrule \hline
    CE & $72.9$ & $83.7$ & $74.8$ & $86.4$ \\  
    \hline
    CE w/ Sub$^\dag$ & $73.5_{(69.2)}$  & $84.6_{(83.1)}$ & $76.2_{(72.3)}$ & $86.7_{(85.5)}$ \\
    SD$^\dag$ & $73.5_{(\mathbf{71.5})}$ & $84.7_{(\mathbf{85.1})}$ & $76.1_{(\mathbf{73.3})}$ & $86.7_{(86.9)}$ \\
    SelfCon$^\dag$ & $74.5_{(70.6)}$ & $84.8_{(84.2)}$ & $76.8_{(72.6)}$ & $87.3_{(85.8)}$ \\
    \hline
    SelfCon & $\mathbf{75.4}_{(69.1)}$ & $\mathbf{86.1}_{(85.2)}$ & $\mathbf{78.5}_{(\mathbf{73.3})}$ & $\mathbf{88.7}_{(\mathbf{87.6})}$ \\ \bottomrule
    \end{tabular}}
\caption{1-stage training on ResNet architectures. $\dag$ describes a modification to 1-stage training with a multi-exit framework. Parentheses indicate the sub-network's classification accuracy. The last row is the results for 2-stage SelfCon. }\label{tab:1-stage}
\end{table}



\section{Ablation Studies}
\label{app:ablation_studies}

\subsection{Different Encoder Architectures}
\label{sec:different_encoder}

We experimented with other architectures: VGG-16 \citep{vgg} with Batch Normalization (BN) \citep{batch_norm} and WRN-16-8 \citep{wrn}, and the results of VGG and WRN are presented in Table \ref{tab:represent_wrn}. The classification accuracy for WRN-16-8 showed a similar trend as that of ResNet architectures. However, for VGG-16 with BN architecture, SupCon had a lower performance than CE on every dataset. Although the contrastive learning approach does not seem to result in significant changes for the VGG-16 with BN encoders, SelfCon was better than or comparable to CE. 

We have also experimented with even more lightweight network because prior works argued that the contrastive learning poorly performs in lightweight architectures \citep{seed, gao2021disco}. For this, we used EfficientNet-b0 \citep{efficientnet} on CIFAR-100, and for SelfCon we used the \textit{fc} sub-network attached at the end of the $4^{\text{th}}$ block. The results are 75.81\%\,(CE), 76.06\%\,(SupCon), and 77.96\%\,(SelfCon), which implies the superior performance of SelfCon learning.

\begin{table*}[!h]
\small
\centering
\vspace{5pt}
\resizebox{0.97\textwidth}{!}{
\renewcommand{\arraystretch}{1.2}
\begin{tabular}{lccccccccc}
\toprule
                   &        &             & \multicolumn{3}{c}{WRN-16-8} & \multicolumn{3}{c}{VGG-16 wih BN}\\ \cmidrule(l{2pt}r{2pt}){4-6} \cmidrule(l{2pt}r{2pt}){7-9}
\multirow{-2.5}{*}{Method} &  \multirow{-2.5}{*}{\thead{Single-\\View}} & \multirow{-2.5}{*}{\thead{Sub-\\Network}} &   CIFAR-10  & CIFAR-100    &   Tiny-ImageNet  &   CIFAR-10  & CIFAR-100    &   Tiny-ImageNet   \\ \midrule \hline
       CE      &       \checkmark &   &  $94.6_{\pm{0.1}}$   &  $73.6_{\pm{0.6}}$   &   $56.5_{\pm{0.5}}$ &  $\mathbf{93.8_{\pm{0.3}}}$    &    $71.2_{\pm{0.2}}$  & $\mathbf{60.7_{\pm{0.1}}}$   \\ \hline
    SupCon &    & &   $95.3_{\pm{0.0}}$    & $75.1_{\pm{0.3}}$      &  $57.4_{\pm{0.3}}$  &  $\mathbf{93.6_{\pm{0.1}}}$    &   $69.6_{\pm{0.1}}$      & $57.3_{\pm{0.4}}$  \\
   SelfCon-M     &   &  \checkmark  &  $\mathbf{95.4_{\pm{0.2}}}$   &      $75.6_{\pm{0.1}}$ & $58.7_{\pm{0.1}}$ &  $93.4_{\pm{0.1}}$     &   $71.7_{\pm{0.3}}$      & $59.4_{\pm{0.1}}$    \\
    SupCon-S &    \checkmark  & &   $95.2_{\pm{0.1}}$   &  $76.0_{\pm{0.1}}$    & $57.3_{\pm{0.5}}$  &  $\mathbf{93.8_{\pm{0.3}}}$       &   $71.1_{\pm{0.0}}$      & $58.4_{\pm{0.2}}$ \\
   \textbf{SelfCon}   &    \checkmark & \checkmark &   $\mathbf{95.5_{\pm{0.0}}}$    &    $\mathbf{76.6_{\pm{0.1}}}$  &    $\mathbf{59.3_{\pm{0.2}}}$   &  $\mathbf{93.5_{\pm{0.1}}}$      &   $\mathbf{72.0_{\pm{0.0}}}$     & $\mathbf{60.7_{\pm{0.1}}}$   \\ \bottomrule
\end{tabular}}
\caption{ The results of linear evaluation on WRN-16-8 and VGG-16 with BN for various datasets. We tuned the best structure and position of the sub-network for each architecture. Appendix \ref{app:ablation_subnet} summarizes the implementation details.}
\label{tab:represent_wrn}
\end{table*}



\subsection{ImageNet with Smaller Batch Size}\label{app:imagenet_small_batch}

We summarized the full results of the ImageNet benchmark with 1024 batch size in Table \ref{tab:ablation_batch_aug} (Left). 
For smaller batch sizes, the multi-viewed methods\,(SupCon and SelfCon-M) outperformed their single-viewed counterparts\,(SupCon-S and SelfCon). SupCon even showed better performance than SelfCon, and SupCon-S showed lower accuracy than CE. We suppose that ImageNet on ResNet-18 can cause an underfitting problem due to the relatively large sample size compared to the small architecture. Moreover, small batch size makes large randomness, as described in Section \ref{exp:single_view}.
In Table \ref{tab:imagenet_total}, as the batch size increases to 2048, the performance of single-viewed methods has increased significantly. 
In particular, SelfCon outperformed SupCon in $B=2048$, which is consistent with the overall experimental results. However, SelfCon-M still achieved slightly higher accuracy than SelfCon, which reflects the need for a deep architecture on large-scale benchmarks.


\subsection{Different Augmentation Policies}
\label{sec:augmentation_policies}

Multi-viewed methods have a problem that oracle needs to choose the augmentation policies carefully \citep{cmc, simclr, swav, mixco}. However, it is difficult and time-consuming to find the optimal policy. We investigated the following augmentation policies to claim that SelfCon reduces the burden of optimizing the augmentation policy.

\begin{itemize}[leftmargin=*]
    \item \textbf{Standard}: For standard augmentation, we used \{RandomResizedCrop, RandomHorizontalFlip, RandomColorJitter, RandomGrayscale\}. This is the same basic policy we used in the paper. 
    \item \textbf{Simple}: When we do not have domain knowledge, it might be difficult to choose the appropriate augmentation policies. We assumed a scenario where we might not know that the color would be important in this visual recognition task. Therefore, we removed the color-related augmentation policies from \textbf{Standard} policy, i.e., we only used \{RandomResizedCrop, RandomHorizontalFlip\} for a simple augmentation policy.
    \item \textbf{RandAugment} : We used RandAugment \citep{cubuk2020randaugment} for an augmentation policy. RandAugment randomly samples $N$ out of 14 transformation choices (e.g., shear, translate, autoContrast, and posterize) with $M$ magnitude parameter. We used the optimized value of $N=2$ and $M=9$ in \citep{cubuk2020randaugment}. It is already known that SupCon performs best with RandAugment policy \citep{supcon}.
\end{itemize}



The results are presented in Table \ref{tab:ablation_batch_aug} (Right). When we apply \textbf{Standard} and \textbf{Simple} augmentations, SelfCon still outperformed SupCon. It supports that SelfCon learning is a more efficient algorithm because finding the best policy, such as RandAugment \citep{cubuk2020randaugment} or AutoAugment \citep{autoaugment}, is not a trivial process and needs a lot of computational costs. Meanwhile, SupCon with the multi-viewed batch can benefit more from the strong and optimized augmentation policy since training each sample twice more encourages memorization. SelfCon learning did not work well with RandAugment, as SupCon degraded with the Stacked RandAugment \citep{infomin} in their experiments, but there would also be an optimal policy for SelfCon. We leave the experiments with other benchmarks, architectures, and various augmentation policies as future work.

\begin{table}[!t]
    \begin{subfigure}[b]{0.4\linewidth}
    \raggedleft
    \small
    \begin{tabular}{lccc}
    \toprule
    Method & Mem. & Time. & Acc@1 \\ \midrule \hline
    CE  & - & - & $69.4$  \\
    \hline
    SupCon & $\times 1.6$ & $\times 2.1$ & $70.9$ \\
    SelfCon-M & $\times 1.5$ & $\times 2.1$ & $\mathbf{71.2}$ \\
    SupCon-S & $\times \mathbf{1.0}$ & $\times \mathbf{1.0}$ & $69.2$ \\
    SelfCon & \underline{$\times 1.0$} & \underline{$\times \mathbf{1.0}$} & $70.3$ \\ \bottomrule
    \end{tabular}
    \end{subfigure}
    \hfill
    \begin{subfigure}[b]{0.5\linewidth}
    \raggedright
    \small
    \begin{tabular}{lccc}
    \toprule
    & \multicolumn{3}{c}{Augmentation Policy} \\
    \cmidrule(l{2pt}r{2pt}){2-4}
    \multirow{-2.5}{*}{Method} & Standard & Simple & RandAugment \\ \midrule \hline
    SupCon & $73.0_{\pm 0.0}$ & $72.0_{\pm 0.3}$ & $\mathbf{74.3_{\pm 0.1}}$ \\
    SelfCon &  $\mathbf{75.4_{\pm 0.1}}$ & $\mathbf{74.2_{\pm 0.2}}$ & $72.5_{\pm 0.0}$ \\ \bottomrule
    \end{tabular}
    \end{subfigure}
    \caption{(Left) The classification accuracy for ImgaeNet with 1024 batch size. We used a ResNet-18 backbone network. \\(Right) CIFAR-100 results on ResNet-18 with various augmentation policies.}
    \label{tab:ablation_batch_aug}
\end{table}

\clearpage
\section{Ensemble Performance of SelfCon in ResNet-18}\label{app:ensemble_resnet50}

We proposed the ensemble of the backbone and sub-network as the advantage of the SelfCon learning framework. Ensembling the multiple outputs is still efficient because the additional fine-tuning of a linear classifier after the frozen sub-network does not place a lot of burden on the computational cost. We summarized the experimental results for ResNet-50 in Table \ref{tab:subnet_ensemble} and ResNet-18 in Table \ref{tab:subnet_ensemble_resnet50}. Ensembling improved the performance by a large margin in every encoder architecture and benchmark. In addition, some of the ensemble results of ResNet-18 significantly outperformed SupCon on ResNet-50: for CIFAR-100, 77.4\% (SelfCon ensemble on ResNet-18) vs.\,75.5\% (SupCon on ResNet-50), and for Tiny-ImageNet, 62.2\% (SelfCon ensemble on ResNet-18) vs.\,61.6\% (SupCon on ResNet-50).

\begin{table*}[!ht]
    \centering
    \small
    \vspace{5pt}
    \begin{tabular}{lccccc}
    \toprule
    Method & CF-10 & CF-100 & Tiny-IN & IN-100 & IN  \\ \midrule \midrule
    SupCon & $94.7$ & $73.0$ & $56.9$ & $85.6$ & $71.2$ \\ \midrule
    Backbone & $\mathbf{95.3}$  & $75.4$ &  $59.8$ & $86.1$ & $71.4$ \\
    Sub-network & $92.6$  & $69.1$ & $53.5$ & $85.2$ & $71.3$  \\
    Ensemble & $95.2$ & $\mathbf{77.4}$ &  $\mathbf{62.2}$ & $\mathbf{87.1}$ & $\mathbf{72.6}$  \\
    \bottomrule
    \end{tabular}
    \caption{Classification accuracy with the classifiers after backbone, sub-network, and the ensemble of them. The ResNet-18 encoder is pretrained by the SelfCon loss function.}
    \label{tab:subnet_ensemble_resnet50}
\end{table*}

\section{Memory Usage and Computational Cost}
\label{app:memory}

We reported the detailed computational cost for pretraining with ResNet-18 and ResNet-50 in Table \ref{tab:memory_flops} and Table \ref{tab:memory_flops_resnet50}, respectively.
All numbers for the ImageNet-100 and ImageNet results are measured on 8 RTX A5000 GPUs, and in other benchmarks on 8 RTX 2080 Ti GPUs. The overall trend for each backbone network is similar.
Although SelfCon has larger parameters due to the auxiliary networks, the actual training memory and time were lower than SupCon because of the single-viewed batch. In other words, SelfCon with $B=512$ requires a comparable memory cost with SupCon with $B=256$.

\begin{table*}[!ht]
\small
\centering
\vspace{10pt}
\resizebox{0.9\linewidth}{!}{
\renewcommand{\arraystretch}{1.2}
\begin{tabular}{llcccccccc}
\toprule
\multirow{2.5}{*}{Dataset (Image size)}&\multirow{2.5}{*}{Method}& \multirow{2.5}{*}{Params} & \multirow{2.5}{*}{FLOPS} & \multicolumn{2}{c}{$B=256$} & \multicolumn{2}{c}{$B=512$} & \multicolumn{2}{c}{$B=1024$} \\
\cmidrule(l{2pt}r{2pt}){5-6}\cmidrule(l{2pt}r{2pt}){7-8}\cmidrule(l{2pt}r{2pt}){9-10}
& &  &  & Memory & Time & Memory & Time & Memory & Time\\ \midrule \hline
\multirow{2}{*}{CIFAR-100 (32x32)} & SupCon & $\mathbf{11.50}$ \textbf{M} & $1.11$ G & $2.14$ & $\mathbf{0.13}$ & $2.35$ & $0.16$ & $3.18$ & $0.27$ \\
& SelfCon & $11.89$ M & $\mathbf{0.56}$ \textbf{G} & $\mathbf{1.83}$ & $\mathbf{0.13}$ & $\mathbf{2.03}$ & $\mathbf{0.14}$ & $\mathbf{2.54}$ & $\mathbf{0.18}$ \\
\hline
\multirow{2}{*}{Tiny-ImageNet (64x64)} & SupCon & $\mathbf{11.50}$ \textbf{M} & $1.13$ G & $2.01$ & $0.14$ & $2.69$ & $0.17$ & $3.97$ & $0.31$ \\
& SelfCon & $11.89$ M & $\mathbf{0.56}$ \textbf{G} & $\mathbf{1.75}$ & $\mathbf{0.13}$ & $\mathbf{2.05}$ & $\mathbf{0.13}$ & $\mathbf{2.68}$ & $\mathbf{0.18}$ \\
\hline
\multirow{2}{*}{ImageNet-100 (224x224)} & SupCon & $\mathbf{11.50}$ \textbf{M} & $3.64$ G & $3.34$ & $0.51$ & $5.34$ & $1.04$ & $9.54$ & $2.11$ \\
& SelfCon & $16.43$ M & $\mathbf{2.18}$ \textbf{G} & $\mathbf{2.54}$ & $\mathbf{0.35}$ & $\mathbf{3.38}$ & $\mathbf{0.70}$ & $\mathbf{5.67}$ & $\mathbf{1.38}$ \\
\bottomrule
\end{tabular}}
\caption{Memory (GiB / GPU) and computation time (sec / step) comparison. All numbers are measured with ResNet-18. Note that FLOPS is for one sample. $B$ stands for batch size. 
}
\label{tab:memory_flops}
\end{table*}

\begin{table}[!ht]
\small
\centering
\resizebox{0.9\textwidth}{!}{
\renewcommand{\arraystretch}{1.05}
\begin{tabular}{llcccccccc}
\toprule
\multirow{2.5}{*}{Dataset (Image size)}&\multirow{2.5}{*}{Method}& \multirow{2.5}{*}{Params} & \multirow{2.5}{*}{FLOPS} & \multicolumn{2}{c}{$B=256$} & \multicolumn{2}{c}{$B=512$} & \multicolumn{2}{c}{$B=1024$} \\
\cmidrule(l{2pt}r{2pt}){5-6}\cmidrule(l{2pt}r{2pt}){7-8}\cmidrule(l{2pt}r{2pt}){9-10}
& &  &  & Memory & Time & Memory & Time & Memory & Time\\ \midrule \hline
\multirow{2}{*}{CIFAR-100 (32x32)} & SupCon & \textbf{27.96 M} & 2.62 G & 4.00 & \textbf{0.28} & 6.40 & 0.35 & 11.29 & 0.50 \\
& SelfCon & 33.47 M & \textbf{1.31 G} & \textbf{2.73} & \textbf{0.28} & \textbf{3.92} & \textbf{0.31} & \textbf{6.28} & \textbf{0.40} \\
\hline
\multirow{2}{*}{Tiny-ImageNet (64x64)} & SupCon & \textbf{27.96 M} & 2.63 G & 4.41 & \textbf{0.27} & 6.71 & 0.33 & 11.84 & 0.46 \\
& SelfCon & 33.47 M & \textbf{1.32 G} & \textbf{2.98} & \textbf{0.27} & \textbf{4.21} & \textbf{0.29} & \textbf{6.82} & \textbf{0.34} \\
\hline
\multirow{2}{*}{ImageNet-100 (224x224)} & SupCon & \textbf{27.97 M} & 8.28 G & 9.41 & 0.61 & 16.49 & 1.14 & - & - \\
& SelfCon & 42.21 M & \textbf{5.33 G} & \textbf{5.91} & \textbf{0.47} & \textbf{10.45} & \textbf{0.72} & - & - \\
\bottomrule
\end{tabular}}
\caption{Memory (GiB / GPU) and computation time (sec / step) comparison. All numbers are measured with ResNet-50. Note that FLOPS is for one sample. $B$ stands for batch size. The 1024 batch size results on the ImageNet-100 benchmark are not reported because ResNet-50 with batch size over 1024 exceeded the GPU limit.}
\label{tab:memory_flops_resnet50}
\end{table}

\clearpage
\section{Correlation Between SelfCon Loss and the MI Estimation}
\label{app:mi_details}


We used three types of MI estimators: InfoNCE \citep{cpc}, MINE \citep{mine}, and NWJ \citep{nwj}. Specifically, we extracted the features of the CIFAR-100 dataset from the pretrained ResNet-18 encoders and optimized a simple 3-layer Conv-ReLU network with the MI estimator objectives. 

In Section \ref{exp:mi_estimation}, to clearly show the correlation between the mutual information and classification accuracy, we experimented with the interpolation between SupCon loss and SelfCon-M loss (SupCon loss is a special case of SelfCon-M loss). However, the current formulation of Eq. \ref{eq:supcon} and Eq. \ref{eq:self-contrastive loss} cannot make the exact interpolation between SupCon and SelfCon-M because the SelfCon-M loss should have negative pairs from different levels of a network (i.e., backbone and sub-network), but the SupCon loss cannot produce those. Therefore, we proposed a near-interpolated loss function between SupCon and SelfCon-M loss, which is equivalent to the supervised version of Eq. \ref{eq:selfcon-mu without fg}.

\paragraph{Loss function}

\begin{align}
     &\gL_{self\text{-}m*} = \frac{1}{1+\alpha}\,\mathcal{L}_{sup} \,+ \nonumber\\
     &\frac{\alpha}{1+\alpha} \Bigg[ - \sum_{\textcolor{anchor}{i} \in I} \frac{1}{|P_{i1}||\bm{\Omega}|} \sum_{\substack{\textcolor{pos}{p_1} \in P_{i1},\\\bm{\omega}_1\in \bm{\Omega}}} \!\log \frac {\exp(\mG(\textcolor{anchor}{\vx_i})^\top \bm{\omega}_1(\textcolor{pos}{\vx_{p_1}}))}{\!\!\sum\limits_{\bm{\omega}_2 \in \bm{\Omega}}\!\!\bigg( \sum\limits_{\textcolor{pos}{p_2} \in P_{i2}} \!\! \exp(\mG(\textcolor{anchor}{\vx_i})^\top \bm{\omega}_2(\textcolor{pos}{\vx_{p_2}})) +\!\! \sum\limits_{\textcolor{neg}{n} \in N_i}\!\! \exp(\mG(\textcolor{anchor}{\vx_i})^\top \bm{\omega}_2(\textcolor{neg}{\vx_n}))\bigg)} \Bigg] \nonumber \\
     \label{eq:selfcon-m_interpolate} 
     &\hspace{40pt}\boxed{I \equiv \{1,\dots,2B\}} \,\,\,
    \boxed{P_{ij} \equiv \{ p_j \in I \setminus \{i\} | y_p=y_i \}} \,\,\,
    \boxed{N_i \equiv \{ n \in I | y_n \neq y_i \}}
\end{align}
\vspace{5pt}

\noindent where $P_{ij} \leftarrow P_{ij} \cup \{i\}$ when $\omega_j = \mF$. Therefore, if $\alpha=0$, $\gL_{self\text{-}m*}$ is equivalent to the SupCon loss and if $\alpha=1$, $\gL_{self\text{-}m*}$ is almost equivalent to SelfCon-M loss. 

Figure \ref{fig:mi_connect_correlation} describes the estimated mutual information and its relationship with classification performance via controlling the hyperparameter $\alpha$, and Table \ref{tab:detail_mi_correlation} summarizes the detailed estimation values of the intermediate feature with respect to the input, label, and the last feature. As expected, SelfCon-M and SelfCon loss have larger MI between the intermediate and the last feature of the backbone network than CE and SupCon loss. We observed a clear increasing trend of both MI and test accuracy as the contribution of SelfCon gets larger (i.e., increasing $\alpha$). When we used a fully-connected layer as the sub-network, we confirmed that the accuracy of SelfCon-M* was quickly saturated to SelfCon-M for $\alpha \geq 0.2$. Note that the detailed MI estimation values in Table \ref{tab:detail_mi_correlation} imply the same interpretation as the IB principle.

\begin{table}[!ht]
\small
\centering
\resizebox{\textwidth}{!}{
\renewcommand{\arraystretch}{1.2}
\begin{tabular}{llccccccccc}
\toprule
 &  & CE & SupCon & SelfCon-M* & SelfCon-M* & SelfCon-M* & SelfCon-M* & SelfCon-M* & SelfCon-M & SelfCon \\ 
\midrule
& $\alpha$ & - & - & $0.025$ & $0.05$ & $0.075$ & $0.1$ & $0.15$ & - & - \\
\midrule
& Accuracy  & $72.9$ & $73.0$ & $73.3$ & $73.5$ & $73.9$ & $74.2$ & $74.6$ & $74.9$ & $\mathbf{75.4}$ \\ 
\midrule \hline
& $\gI(\vx;\mT(\vx))$ & $0.436$ & $0.285$ & $0.296$ & $0.277$ & $0.299$ & $0.293$ & $0.232$ & $\mathbf{0.203}$ & $0.207$ \\ 
& $\gI(\vy;\mT(\vx))$ & $0.233$ & $0.221$ & $0.299$ & $0.290$ & $0.309$ & $0.329$ & $0.341$ & $0.454$  & $\mathbf{0.463}$ \\
\cellcolor{white}  \multirow{-3}{*}{InfoNCE} & $\gI(\mF(\vx);\mT(\vx))$ & $0.313$ & $0.296$ & $0.330$ & $0.357$ & $0.381$  & $0.392$ & $0.402$ & $0.508$ & $\mathbf{0.528}$ \\ \hline
& $\gI(\vx;\mT(\vx))$ & $1.225$ &$0.758$ & $0.843$ & $0.719$ & $0.744$ & $0.665$ & $0.697$ & $0.508$ & $\mathbf{0.503}$ \\
& $\gI(\vy;\mT(\vx))$ & $0.617$ & $0.616$ & $0.719$ & $0.700$ & $0.834$ & $0.928$ & $0.961$ & $1.261$ & $\mathbf{1.425}$ \\
\cellcolor{white} \multirow{-3}{*}{MINE} & $\gI(\mF(\vx);\mT(\vx))$ & $0.786$ & $0.845$ & $0.919$ & $0.937$ & $1.032$ & $1.140$ & $1.050$ & $1.760$ & $\mathbf{2.024}$ \\ 
\hline
& $\gI(\vx;\mT(\vx))$ & $1.201$ & $0.714$ & $0.798$ & $0.694$ & $0.764$ & $0.692$ & $0.592$ & $0.496$ & $\mathbf{0.486}$ \\ 
& $\gI(\vy;\mT(\vx))$ & $0.501$ & $0.467$ & $0.594$ & $0.641$ & $0.808$ & $0.799$ & $0.843$ & $\mathbf{1.287}$ & $1.236$ \\
\cellcolor{white}  \multirow{-3}{*}{NWJ} & $\gI(\mF(\vx);\mT(\vx))$ & $0.778$ & $0.747$ & $0.835$ & $0.914$ & $1.039$ & $1.086$ & $1.101$ & $1.593$ & $\mathbf{1.736}$ \\
 \bottomrule
\end{tabular}
}
\caption{The detailed results of Figure \ref{fig:mi_connect_correlation}. $\vx$, $\vy$, $\mT(\vx)$, and $\mF(\vx)$ respectively denotes the input variable, label variable, intermediate feature, and the last feature. Recall that $\mT(\vx)$ is the intermediate feature of the backbone network, which is an input to the auxiliary network path. We summarized the average of estimated MI through multiple random seeds. 
Bold type indicates the smallest values for $\gI(\vx;\mT(\vx))$ and the largest values for $\gI(\vy;\mT(\vx))$ and $\gI(\mF(\vx);\mT(\vx))$, according to the IB principle. We used ResNet-18 on CIFAR-100 dataset for the measurements.}
\label{tab:detail_mi_correlation}
\end{table}

\clearpage
\section{Qualitative Examples for Vanishing Gradient}

\begin{figure}[!ht]
    \centering
    \includegraphics[width=0.9\textwidth]{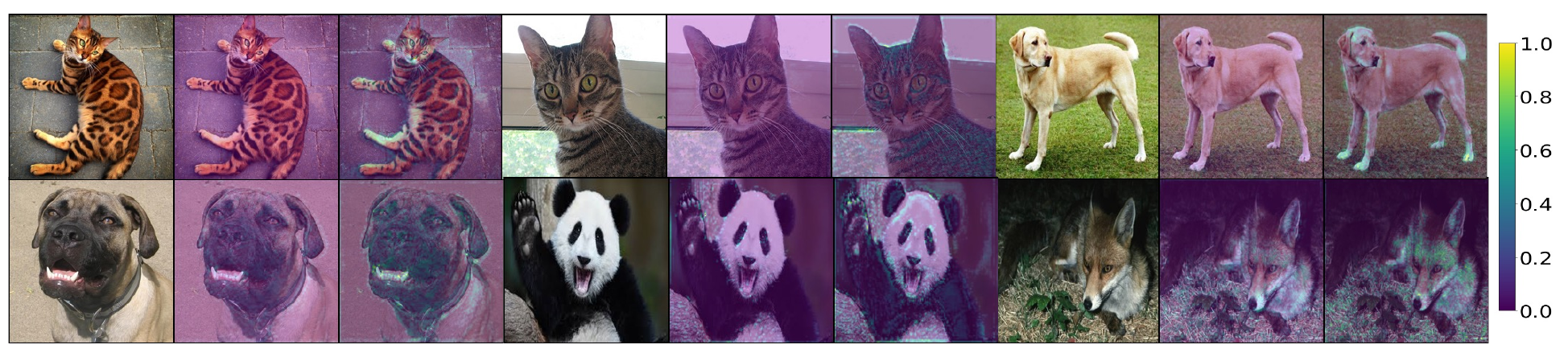}
    \caption{Qualitative examples for mitigating vanishing gradient. Along with the original image, we visualized the gradient when training with SupCon (Left) and SelfCon-M loss (Right). Note that all the gradients are from the same model checkpoint of ResNet-18.}
    \label{fig:qualitative_vis}
\end{figure}

In Figure \ref{fig:vanishing_grad}, we have already shown that the sub-network solves the vanishing gradient problem through the visualization for gradient norms of each layer.
In Figure \ref{fig:qualitative_vis}, we also visualized qualitative examples using Grad-CAM \citep{gradcam}.
We used the gradient measured on the last layer in the $2^\text{nd}$ block when the sub-network is attached after the $2^\text{nd}$ block.
In order to compare the absolute magnitude of the gradient, it is normalized by the maximum and minimum values of the two methods, SelfCon-M and SupCon.
As in Figure \ref{fig:qualitative_vis}, SelfCon learning led to a larger gradient via sub-networks, and Grad-CAM more clearly highlighted the pixels containing important information in the images.

\section{Robustness to Image Corruptions}

SelfCon is superior to SupCon in the perspective of robustness to corrupted images. We followed the same evaluation protocol as in \citet{supcon} that measures Mean Corruption Error\,(mCE) and Relative mCE metrics, averaged over 15 corruptions and 5 severity levels of ImageNet-C dataset \citep{hendrycks2019benchmarking}. For the pretrained ResNet-34 with SelfCon and SupCon, we obtained mCE values of 74.9 and 77.3 and relative mCE values of 99.7 and 102.8 (lower is better), respectively.
As robustness can be achieved by generalization to the corrupted distributions, the result is consistent with our extensive experimental analyses, such as Figure \ref{fig:generalization_error}.
Note that they are normalized values by AlexNet results, as in the original ImageNet-C paper, and relative mCE metric measures the relative error rate compared to the clean data error.

\end{document}